# Multi-View Fuzzy Logic System with the Cooperation between Visible and Hidden Views

Te Zhang, Zhaohong Deng, *Senior Member, IEEE*, Dongrui Wu, *Senior Member, IEEE*, and Shitong Wang

***Abstract*--Multi-view datasets are frequently encountered in learning tasks, such as web data mining and multimedia information analysis. Given a multi-view dataset, traditional learning algorithms usually decompose it into several single-view datasets, from each of which a single-view model is learned. In contrast, a multi-view learning algorithm can achieve better performance by cooperative learning on the multi-view data. However, existing multi-view approaches mainly focus on the views that are visible and ignore the hidden information behind the visible views, which usually contains some intrinsic information of the multi-view data, or vice versa. To address this problem, this paper proposes a multi-view fuzzy logic system, which utilizes both the hidden information shared by the multiple visible views and the information of each visible view. Extensive experiments were conducted to validate its effectiveness.**

## I. Introduction

Multi-view learning is a machine learning paradigm which focuses on the data represented by different feature sets. The main reason for the arising of multi-view learning is that more and more datasets can be represented by different attribute sets, i.e., they are collected with multiple views in practical scenes. For example, during food fermentation process, the food fermentation index can be recorded from two aspects: fermentation conditions and chemical index [3]. Another example is content based web image retrieval, where each image can be described by visible pixels and the associated text simultaneously [1]. Multi-view learning has shown very promising results in such applications [2-7].

Multi-view learning usually aims to learn one model by integrating the information of multiple views to improve the generalization performance. A naive solution for multi-view learning concatenates all views into one single view and applies traditional single-view learning algorithms directly. However, this can easily result in over-fitting when the training set is small, and also the specific statistical property of each view is lost. A noteworthy merit of multi-view learning is that the performance on a single view dataset may still be improved by using artificially generated multiple views.

Many multi-view learning approaches have been proposed in the literature. Representative ones are reviewed below.

*1) Multi-view classification*, which applies multi-view learning to classification problems. A frequently used technique is co-regularization, which adds regularization terms to the objective function to make data from multiple views consistent. For example, an approach combining kernel canonical component analysis and support vector machine (SVM) was proposed in [7], a multi-view transductive SVM was proposed in [5] by introducing global constraint variables, and a multi-view Laplacian SVM was proposed in [6] by integrating manifold regularization and multi-view regularization with a traditional SVM. Ensemble learning has also been extended to multi-view settings [37, 43], and shown good performance on high dimensional data and poem data. A multi-view perceptron using a deep model for learning face identity and view representation was proposed in [38].

*2) Multi-view clustering*, which is unsupervised. A collaborative clustering algorithm based on the classical fuzzy *c*-means (FCM) approach was proposed in [9], which used collaborative partition by controlling the fuzzy partitions among different views. Two different collaborative multi-view clustering approaches were proposed in [2] using more sophisticated view corporation mechanisms. Two new multi-view clustering algorithms based on kernel *k*-means and spectral clustering were proposed in [10]. An expectation-maximization based collaborative multi-view clustering method, Co-EM, was proposed in [11]. A canonical correlation analysis based multi-view clustering technique was proposed in [12]. A novel tensor-based framework for integrating heterogeneous multi-view data in the context of spectral clustering was proposed in [39]. A new robust large-scale multi-view clustering method which integrates multiple representations of large scale data was proposed in [40]. A multi-view learning model, which integrates all features and learns the weight for every feature with respect to each cluster individually, was proposed in [41]. Most of the above methods are based on feature transformation [39-41], which transforms features form different views to a common feature space.

*3) Multi-view regression*. A multi-view regression approach using canonical correlation analysis was proposed in [13]. A multi-view low-rank model which imposes low-rank constraints on the multi-view regression model was proposed

This work was supported in part by the Outstanding Youth Fund of Jiangsu Province (BK20140001), by the National Key Research Program of China under grant 2016YFB0800803, the National Natural Science Foundation of China (61272210). (Corresponding author: Zhaohong Deng)

T. Zhang, Z.H. Deng and S.T. Wang are with the School of Digital Media, Jiangnan University and Jiangsu Key Laboratory of Digital Design and Software Technology, Wuxi 214122, China (e-mail: 511865360 @qq.com; dengzhaohong@jiangnan.edu.cn; wxwangst@aliyun.com)

D. Wu is with the Key Laboratory of the Ministry of Education for Image Processing and Intelligent Control, School of Automation, Huazhong University of Science and Technology, Wuhan, China (e-mail: drwu@hust.edu.cn)

in [35]. A multi-view regression model combining feature selection, low-rank and subspace learning was proposed in [36].

Fuzzy sets and fuzzy logic systems (FLSs) have also been successfully used in multi-view learning. For multi-view clustering, a weighed fuzzy multi-view collaborative clustering algorithm based FCM was proposed in [14] by introducing a penalty factor, and a minimax FCM multi-view clustering algorithm was proposed in [15] by introducing minimax optimization into the classical FCM. For multi-view classification, the large margin learning mechanism was introduced into the objective function of the classical Takagi-Sugeno-Kang (TSK) FLS, and then a collaborative learning based two-view fuzzy classification system was proposed in [16]. A multi-view FLS for epilepsy EEG signal recognition was proposed in [17], which achieved better performance in epilepsy detection than many other methods.

Although existing multi-learning techniques have shown promising performance in different applications, there is still room to improve due to the following fact. Most multi-view data come from different domains or different feature extraction methods. Thus, there exists otherness among different views. Meanwhile, the features come from different views describe the same objects, so there also exists internal relation among different views. Thus multi-view learning usually follows two principles: 1) complementarity principle, and 2) consistency principle. The complementarity principle aims to make full use of the otherness in different views to describe the data in a comprehensive way and enhance the model generalization ability. The consistency principle aims to use the internal relation between different views to maximize the consistency and enhance the model performance. However, most existing methods only focus on the information from the visible views, i.e., the otherness among different views, or only use the internal relation among them. In practice, not only the otherness but also the internal relation among different views are important for multi-view modelling tasks. How to effectively integrate the complementarity principle with consistency principle in multi-view modelling is a challenging task.

A novel multi-view fuzzy classification system, i.e., TSK FLS with cooperation between visible and hidden views (TSK-FLS-CVH), is proposed in this paper. First, a TSK FLS is used as the base model to construct a multi-view fuzzy classification system due to its flexibility and simplicity [18]. Then, non-negative matrix factorization (NMF) is used to learn the hidden space that is shared among the visible views, and the corresponding data are obtained in this hidden view. The NMF based hidden view learning can effectively extract the consistency of different visible views. Next, cooperative learning between the visible and hidden views is implemented to construct the multi-view TSK FLS. By cooperative learning, the visible views and the shared hidden view can benefit each other according to the complementarity principle. Compared with existing multi-view FLSs and other related multi-view approaches, the proposed method can effectively use the shared hidden information to implement the cooperation between visible and hidden views, i.e., it can effectively use the otherness and consistency in multi-view data, which greatly enhances the generalization performance of the trained model.

The main contributions of this paper can be highlighted as follows:

*1)* An approach to extract shared hidden information among the visible views is proposed using non-negative matrix factorization.

*2)* A multi-view FLS with cooperation between the visible and hidden views is proposed.

*3)* The proposed TSK-FLS-CVH is validated using extensive experiments.

The rest of this paper is organized as follows. Section II briefly reviews the concepts and principles of classical TSK FLSs and multi-view learning. Section III proposes a strategy of learning the shared hidden space from multiple visible views, and describes the details of the multi-view FLS, TSK-FLS-CVH. Section IV presents the experimental results. Finally, Section V draws conclusions and points out some future research directions.

## II. Background Knowledge

The basic concepts and principles of classical TSK FLSs and multi-view learning are briefly reviewed in this section.

### *A. TSK FLS*

There are three main categories of FLSs: TSK Model [18], Mamdani Model [19], and Generalized Fuzzy Model [20]. Among them, the TSK model is most popular due to its simplicity and flexibility, and it is used in this paper.

#### *1) TSK Fuzzy Classification System*

A TSK FLS employs the following rules:

$$\begin{aligned}&\mathrm{R}^k\text{: IF } x_1 \text{ is } A_1^k \wedge x_2 \text{ is } A_2^k \wedge \ldots \wedge x_d \text{ is } A_d^k\\&\text{THEN } f^k(x) = p_0^k + p_1^k x_1 + \ldots + p_d^k x_d,\; k = 1,2,\ldots,K\end{aligned} \tag{1}$$

where $\boldsymbol{x} = [x_1, x_2, \ldots, x_d]^{\mathrm{T}}$ is an input vector, $A_i^k$ is a fuzzy set in the $i$th input domain for the $k$th rule, and $\wedge$ is a fuzzy conjunction operator.

When the product $t$-norm and center-of-gravity defuzzification are used, the final output of a classical TSK FLS is calculated as

$$y^0 = \sum_{k=1}^{K} \frac{\mu^k(\boldsymbol{x})}{\sum_{k'}^{K} \mu^{k'}(\boldsymbol{x})} f^k(x) = \sum_{k=1}^{K} \tilde{\mu}^k(\boldsymbol{x}) f^k(\boldsymbol{x}) \tag{2a}$$

where

$$\mu^k(\boldsymbol{x}) = \prod_{i=1}^{d} \mu_{A_i^k}(x_i) \tag{2b}$$

$$\tilde{\mu}^k(\boldsymbol{x}) = \mu^k(\boldsymbol{x}) \Big/ \sum_{k'=1}^{K} \mu^{k'}(\boldsymbol{x}) \tag{2c}$$

in which $\mu_{A_i^k}(x_i)$ is the membership of $x_i$ to fuzzy set $A_i^k$,

$\mu^k(\boldsymbol{x})$ is the firing strength of the $k$th rule to input vector $\boldsymbol{x}$, and $\sum_{k'=1}^{K}\mu^{k'}(\boldsymbol{x})$ is a normalization term which is the sum of all firing strengths of $K$ fuzzy rules to $\boldsymbol{x}$. With the normalization term, the normalized firing strength of the $k$th rule to $\boldsymbol{x}$, i.e., $\tilde{\mu}^k(\boldsymbol{x})$ can be calculated. Gaussian membership functions are used in this paper:

$$\mu_{A_i^k}(x_i) = \exp\left(\frac{-\left(x_i - c_i^k\right)^2}{2\delta_i^k}\right) \quad (2d)$$

where the parameters can be estimated using different approaches, e.g., FCM clustering [1]:

$$c_i^k = \sum_{j=1}^{N} u_{jk} x_{ji} \Big/ \sum_{j=1}^{N} u_{jk} \quad (2e)$$

$$\delta_i^k = h \cdot \sum_{j=1}^{N} u_{jk}\left(x_{ji} - c_i^k\right)^2 \Big/ \sum_{j=1}^{N} u_{jk} \quad (2f)$$

in which $u_{jk}$ denotes the membership of $x_{ji}$ in the $k$th fuzzy partition obtained by FCM on the training dataset [23], and $h$ is a scaling parameter which can be set manually or determined by cross-validation (CV).

When the antecedent parameters in the rules are fixed, a TSK FLS can be re-expressed as a linear model [3] [21] [24]. Let

$$\mathbf{x}_e = \left(1, \mathbf{x}^T\right)^T \quad (3a)$$

$$\tilde{\mathbf{x}}^k = \tilde{\mu}^k(\mathbf{x})\mathbf{x}_e \quad (3b)$$

$$\mathbf{x}_g = \left(\left(\tilde{\mathbf{x}}^1\right)^T, \left(\tilde{\mathbf{x}}^2\right)^T, \ldots, \left(\tilde{\mathbf{x}}^K\right)^T\right)^T \quad (3c)$$

$$\boldsymbol{p}^k = \left(p_0^k, p_1^k, \ldots, p_d^k\right)^T \quad (3d)$$

$$\boldsymbol{p}_g = \left(\left(\boldsymbol{p}^1\right)^T, \left(\boldsymbol{p}^2\right)^T, \ldots, \left(\boldsymbol{p}^K\right)^T\right)^T \quad (3e)$$

Then, (2a) becomes

$$y^0 = \boldsymbol{p}_g^T \mathbf{x}_g \quad (3f)$$

where $\mathbf{x}_g \in R^{K(d+1)}$ is a data vector in the new feature space that is mapped from the data vector $\mathbf{x} \in R^d$ in the original feature space, and $\boldsymbol{p}_g$ is the combination of the consequent parameters of all fuzzy rules, which can be optimized using the method proposed later in this paper.

*2) Classification based on the TSK FLS regression model*

TSK FLSs have been extensively used for regression tasks. A classification task can be transformed into a multi-output regression task, and then a TSK FLS can be applied. For example, a three-class classification problem can be transformed to a three-output regression problem by representing labels 1, 2, 3 as $[1\ 0\ 0]^{\mathrm{T}}$, $[0\ 1\ 0]^{\mathrm{T}}$, and $[0\ 0\ 1]^{\mathrm{T}}$, respectively. The predicted label of a test sample is the index of the maximum value in the output vector. For example, if the prediction output of a test sample is $[0.6\ 0.3\ 0.1]^{\mathrm{T}}$, then the test sample is classified to Class 1.

## B. *Multi-view Learning*

Multi-view learning trains a model by considering the correlation and cooperation among different views, as shown in Fig. 1. Generally the performance of a multi-view learning model is better than each individual single-view model, as more information is exploited in the former.

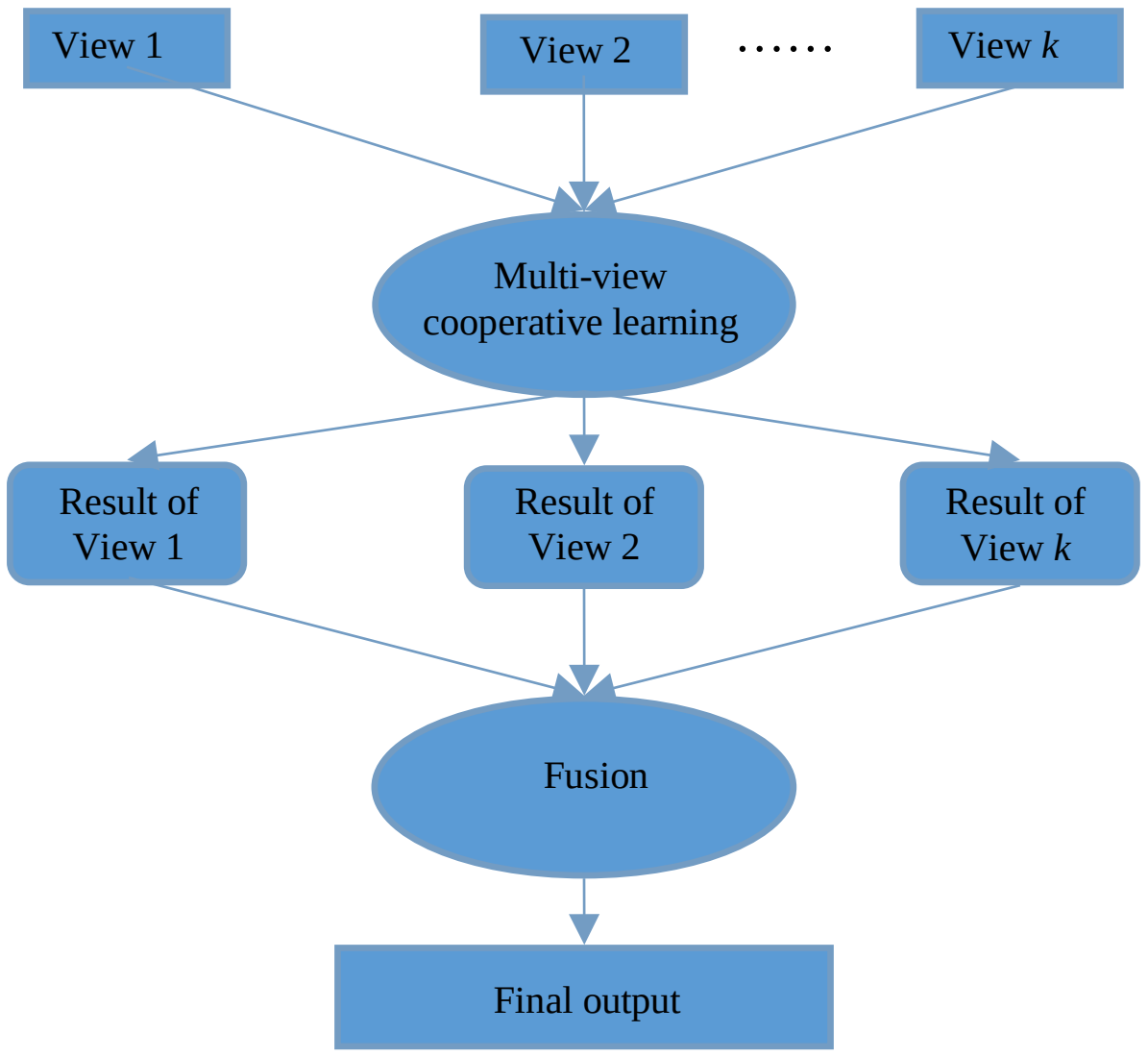

Fig. 1 A classical multi-view learning framework.

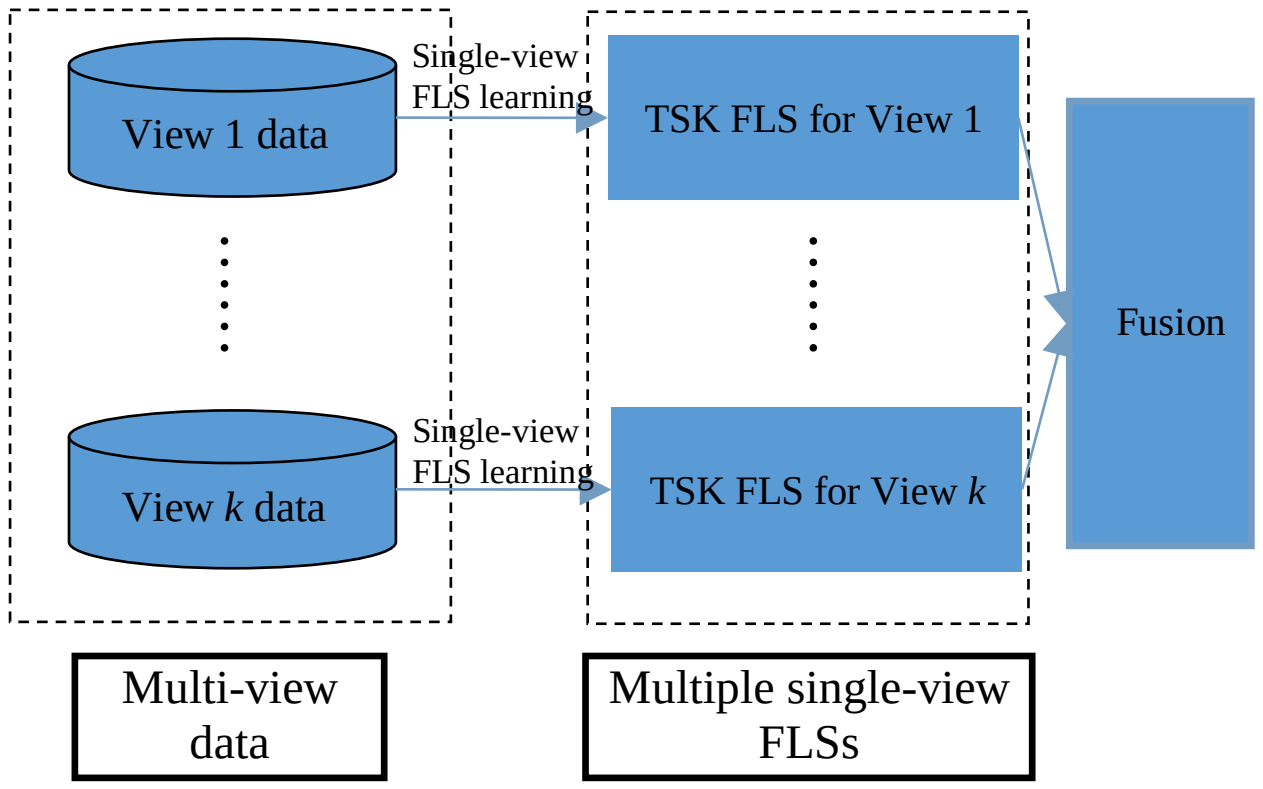

Fig. 2 Multi-view modeling using single-view TSK FLS.

FLSs are frequently used in building individual single-view models. For multi-view data, a popular strategy is to build a single-view FLS for every view and then fuse their outputs, e.g., using averaging, as shown in Fig. 2.

The above simple strategy provides a feasible way for using single-view models for multi-view data, but it ignores the relationship among the multiple views, and hence may not be able to get the best learning performance. Research has been done to improve it. A novel two-view FLS which combines cooperation leaning and large margin criterion was proposed in [16], which demonstrated promising performance. In addition, a multi-view FLS has also been proposed in [17] for epilepsy EEG recognition. In this paper we propose further improvements to multi-view FLSs, by exploiting the hidden information shared among different visible views and using the cooperation between the visible and hidden views.

## III. Multi-view FLS With Cooperation Between Visible and Hidden Views

This section proposes TSK-FLS-CVH (Fig. 3), a novel multi-view FLS with the cooperation between the visible and hidden views. It uses multi-view cooperation to train multiple FLSs for different views. The cooperation not only exists among the visible views, but also exists between the visible and hidden views. In this way, it can simultaneously utilize the explicit information from the visible views and the shared hidden information among them.

The details of TSK-FLS-CVH are described next.

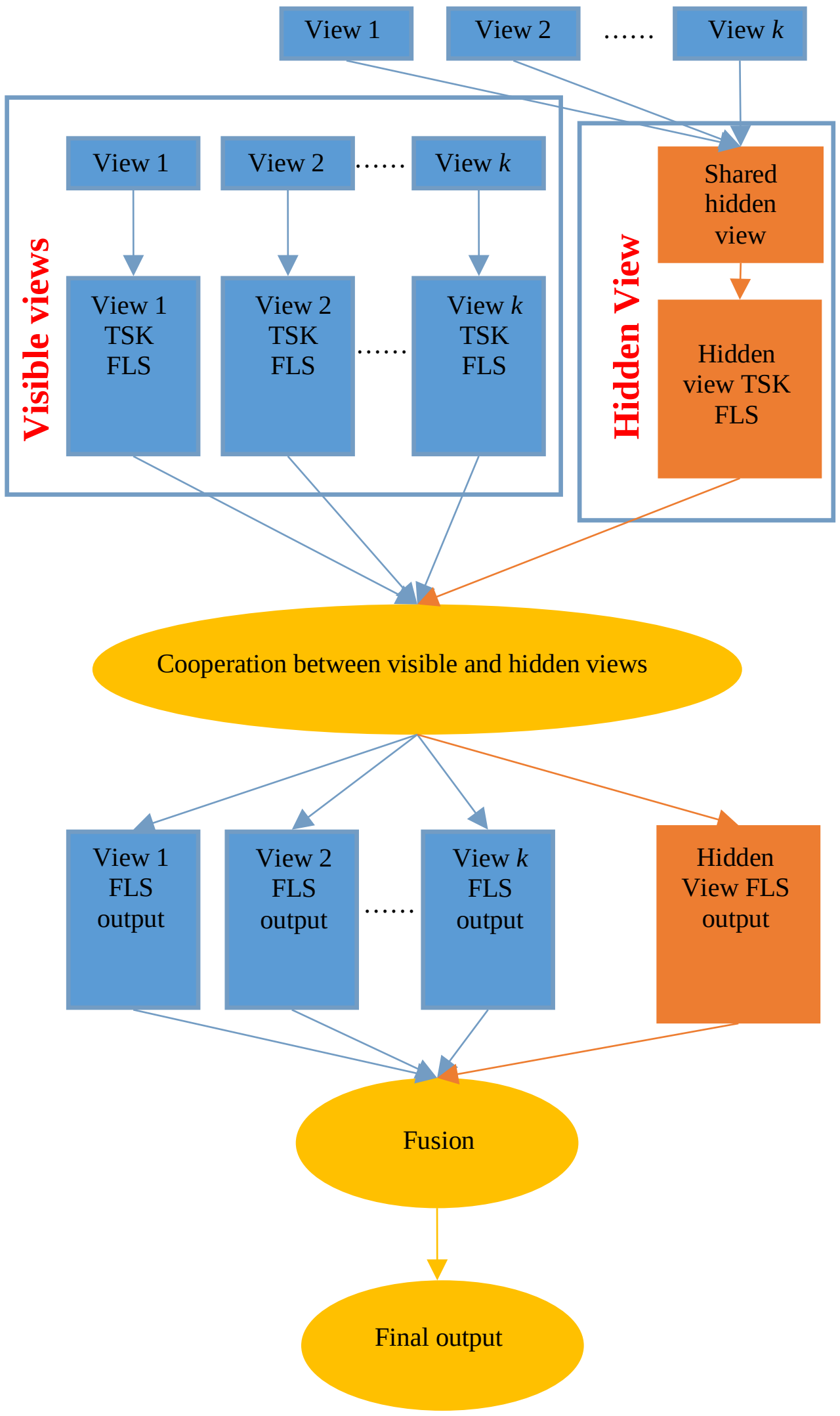

Fig. 3 The framework of TSK-FLS-CVH.

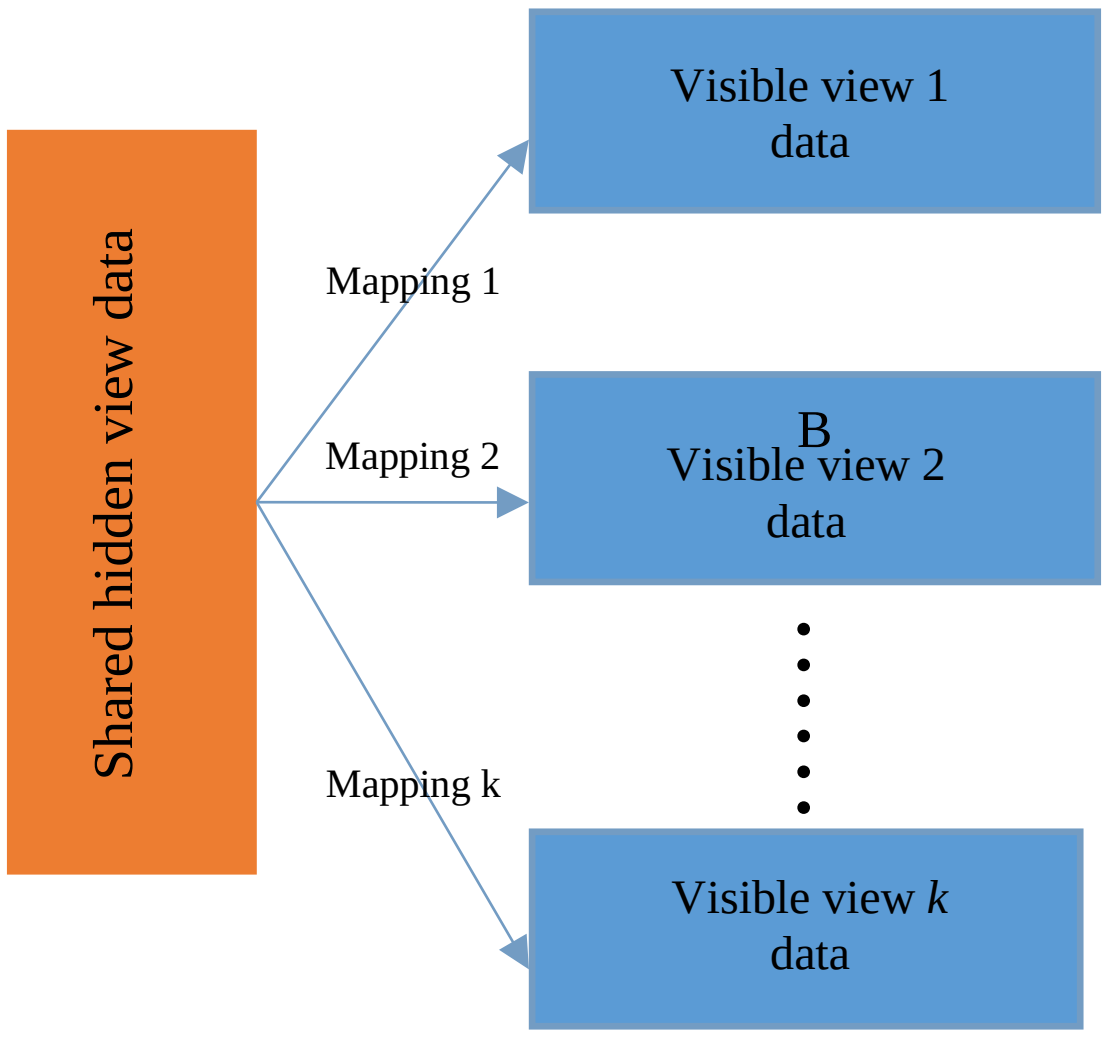

Fig.4 The relationship between the hidden view data and the visible view data.

### A. *Hidden View Generation*

Our hypothesis is that there exists a shared hidden space that can be used to generate different visible spaces by different mappings, as shown in Fig. 4. This subsection describes how to identify this hidden view.

Given a multi-view dataset with $N$ samples and $K$ visible views, the feature subset associated with the $k$th visible view can be represented as $\{\tilde{\boldsymbol{x}}_i^k\}_{i=1}^N \subset R^{d_k}$, and the corresponding matrix representation is $\bar{\boldsymbol{X}}^k = \left[\tilde{\boldsymbol{x}}_1^k, \ldots, \tilde{\boldsymbol{x}}_N^k\right]^T \in \mathrm{R}^{N\times d_k}$. Let $\boldsymbol{H} = [\boldsymbol{h}_1, \boldsymbol{h}_2, \ldots, \boldsymbol{h}_N]^T \in R^{N\times r}$ be the data representation in the shared hidden space, and $\boldsymbol{W}^k \in R^{r\times d_k}$ be the mapping between the hidden view and the $k$th visible view. They are then identified by solving the following minimization problem:

$$\min_{\{\boldsymbol{W}^k\}_{k=1}^K, \boldsymbol{H}} \sum_{k=1}^K \left(\alpha^k l\left(\bar{\boldsymbol{X}}^k, \boldsymbol{W}^k, \boldsymbol{H}\right) + \beta^k Dist\left(\kappa_{\bar{\boldsymbol{X}}^k}, \kappa_{\boldsymbol{H}}\right)\right) \quad (4)$$
$$s.t. \quad \sum_{k=1}^K \alpha^k = 1$$

where $\alpha^k \geq 0$ and $\beta^k \geq 0$ represents the weight and the regularization coefficient of the $k$th visible view, respectively. The first term of (4), $l\left(\bar{\boldsymbol{X}}^k, \boldsymbol{W}^k, \boldsymbol{H}\right)$ is the empirical loss in mapping from $\boldsymbol{H}$ to $\bar{\boldsymbol{X}}^k$, which is defined below,

$$l\left(\bar{\boldsymbol{X}}^k, \boldsymbol{W}^k, \boldsymbol{H}\right) = \left\|\bar{\boldsymbol{X}}^k - \boldsymbol{H}\boldsymbol{W}^k\right\|_F^2 \quad (5)$$
$$s.t. \quad \boldsymbol{W}^k, \boldsymbol{H} \geq \boldsymbol{0}$$

The second term of (4), i.e., $Dist\left(\kappa_{\bar{\boldsymbol{X}}^k}, \kappa_{\boldsymbol{H}}\right)$ is a regularization term. It is developed to guarantee that the data similarity $\kappa_{\boldsymbol{H}}$ described by the hidden view is close to the similarity $\kappa_{\bar{\boldsymbol{X}}^k}$ described by the original view. For this purpose, the manifold regularization is adopted in this paper. First, a nearest neighbor graph with $N$ vertexes $G^k$ for each visible view data is generated. Let $\boldsymbol{S^k} \in R^{N\times N}$ be the weight matrix of $G^k$. If $\tilde{\boldsymbol{x}}_{\boldsymbol{i}}^{\boldsymbol{k}}$ is the $\varepsilon$-nearest neighbor of $\tilde{\boldsymbol{x}}_{\boldsymbol{j}}^{\boldsymbol{k}}$ or $\tilde{\boldsymbol{x}}_{\boldsymbol{j}}^{\boldsymbol{k}}$ is the $\varepsilon$-nearest neighbor of $\tilde{\boldsymbol{x}}_{\boldsymbol{i}}^{\boldsymbol{k}}$ (the value of $\varepsilon$ can be set manually or other strategies), then $S_{ij}^k = \kappa\left(\tilde{\boldsymbol{x}}_i^k, \tilde{\boldsymbol{x}}_j^k\right)$, otherwise, $S_{ij}^k = 0$. $\kappa\left(\tilde{\boldsymbol{x}}_i^k, \tilde{\boldsymbol{x}}_j^k\right)$ is a kernel function. If two points are neighbors in the original feature space, then they are also expected to be neighbors in the hidden space. Let $d_i^k = \sum_{j=1}^{N} S_{i,j}$, $\boldsymbol{D^k} = diag\left(d_1^k,...,d_N^k\right) \in R^{N\times N}$, and then we define the regularization term $Dist\left(\kappa_{\bar{\boldsymbol{X}}^k}, \kappa_{\boldsymbol{H}}\right)$ in (4) as follows [27]:

$$Dist\left(\kappa_{\bar{\boldsymbol{X}}^k}, \kappa_{\boldsymbol{H}}\right) = \frac{1}{2}\sum_{i=1}^{N}\sum_{j=1}^{N} \| \boldsymbol{h_i} - \boldsymbol{h_j} \|^2 S_{i,j}^k = Trace\left(\boldsymbol{H}^{\mathrm{T}}\boldsymbol{L}^k\boldsymbol{H}\right) \tag{6}$$

where $\boldsymbol{L^k} = \boldsymbol{D^k} - \boldsymbol{S^k}$ is a Laplacian matrix.

Combining (5) and (6), (4) can be re-expressed as:

$$\min_{\{\boldsymbol{W}^k\}_{k=1}^K, \boldsymbol{H}} \sum_{k=1}^{K}\left(\alpha^k \left\|\bar{\boldsymbol{X}}^k - \boldsymbol{H}\boldsymbol{W}^k\right\|_F^2 + \beta^k Trace\left(\boldsymbol{H}^{\mathrm{T}}\boldsymbol{L}^k\boldsymbol{H}\right)\right) \tag{7}$$

Next we explain how to solve for $\boldsymbol{H}$ and $\boldsymbol{W}^k$ from (7).

We first make all elements of $\bar{\boldsymbol{X}}^k$ non-negative by normalization. Then, under the constraints $\boldsymbol{W}^k \geq 0$ and $\boldsymbol{H} \geq \boldsymbol{0}$, (7) can be solved by non-negative matrix factorization (NMF). Existing solvers, such as coordinate descent [26], can be used directly. More specifically, (7) can be solved iteratively: (i) Optimize $\boldsymbol{W}^k$ by fixing $\boldsymbol{H}$; and (ii) Optimize $\boldsymbol{H}$ by fixing $\boldsymbol{W}^k$.

(i) Optimize $\boldsymbol{W}^k$ by fixing $\boldsymbol{H}$

When $\boldsymbol{H} = \boldsymbol{H}^{(t)}$ is fixed (where *t* is the current iteration number), the sub-optimization problem of (7) that only involves $\boldsymbol{W}^k$ as the variable can be equivalently formulated as follows:

$$\min_{\boldsymbol{W^k}} \left\|\bar{\boldsymbol{X}}^k - \boldsymbol{H}\boldsymbol{W}^k\right\|_F^2 \quad s.t.\ \boldsymbol{W}^k \geq \boldsymbol{0} \tag{8}$$

And $\boldsymbol{W}^k$ can be updated as [26]:

$$\left(\boldsymbol{W}^k\right)_{i,j}^{(t+1)} \leftarrow \left[\frac{\left((\bar{\boldsymbol{X}}^k)^T\boldsymbol{H}\right)_{i,j}}{\left(\boldsymbol{H}^T\boldsymbol{H}\left(\boldsymbol{W}^k\right)^{(t)}\right)_{i,j}}\right]\left(\boldsymbol{W}^k\right)_{i,j}^{(t)} \tag{9}$$

(ii) Optimize $\boldsymbol{H}$ by fixing $\boldsymbol{W}^k$

When $\boldsymbol{W}^k = \left(\boldsymbol{W}^k\right)^{(t+1)}$ is fixed, the optimization problem that involves $\boldsymbol{H}$ becomes:

$$\min_{H} \sum_{k=1}^{K} \alpha^k \left\|\bar{\boldsymbol{X}}^k - \boldsymbol{H}\boldsymbol{W}^k\right\|_F^2 + \beta^k Trace\left(\boldsymbol{H}^{\mathrm{T}}\boldsymbol{L}^k\boldsymbol{H}\right) \quad s.t.\ \boldsymbol{H} \geq 0 \tag{10}$$

And $\boldsymbol{H}$ can be updated as [26]:

$$\boldsymbol{H}_{i,j}^{(t+1)} \leftarrow \left[\frac{\sum_{k=1}^{K}\alpha^k\left(\bar{\boldsymbol{X}}^k\left(\boldsymbol{W}^k\right)^T\right)_{i,j}}{\sum_{k=1}^{K}\left(\alpha^k\left(\boldsymbol{H}^{(t)}\boldsymbol{W}^k\left(\boldsymbol{W}^k\right)^T\right)_{i,j} + \beta^k\left(\boldsymbol{L}^k\boldsymbol{H}^{(t)}\right)_{i,j}\right)}\right]\boldsymbol{H}_{i,j}^{(t)} \tag{11}$$

Clearly, if the initial $\left(\boldsymbol{W}^k\right)^{(0)}$ and $\boldsymbol{H}^{(0)}$ are non-negative, then $\boldsymbol{W}^k$ and $\boldsymbol{H}$ are guaranteed to be non-negative with the above updating rules.

### *B. Cooperative learning between Visible and Hidden Views*

Given a multi-view dataset with $K$ visible views, we obtain a hidden view using the approach introduced in the previous subsection. Cooperative learning is used to make use of information in these $K+1$ views.

Assume the TSK FLS corresponding to the *k*th visible view has $L_k$ rules. Then, it can be represented as the following linear model:

$$f^k(\boldsymbol{x}^k) = (\boldsymbol{p_g}^k)^T\tilde{\boldsymbol{x}}_g^k \tag{12}$$

where $\boldsymbol{p_g}^k$, $\tilde{\boldsymbol{x}}_g^k \in R^{L_k(d_k+1)}$, $\boldsymbol{p_g}^k$ is the vector consisting of all the consequent parameters of the FLS, $\tilde{\boldsymbol{x}}_g^k$ is the vector in the new feature space which is mapped by fuzzy rules from the input vector $\tilde{\boldsymbol{x}}^k$ in the original feature space of View *k*.

Assume the TSK FLS corresponding to the hidden view has *J* rules. Then, it can be represented as:

$$h(\tilde{\boldsymbol{h}}) = (\boldsymbol{p_g}^{K+1})^T\tilde{\boldsymbol{h}}_g \tag{13}$$

where $\boldsymbol{p_g}^{K+1}$, $\tilde{\boldsymbol{h}}_g \in R^{J(r+1)}$, $\boldsymbol{p_g}^{K+1}$ is the vector consisting of all the consequent parameters, $\tilde{\boldsymbol{h}}_g$ is the vector in the new feature space which is mapped by fuzzy rules from the input vector $\tilde{\boldsymbol{h}}$ in the original feature space of the hidden view.

Based on the above transformations, the following optimization problem is proposed for the multi-view TSK FLS with the cooperation of the visible and hidden views:

$$\begin{aligned}\min_{\boldsymbol{p_g},w} J = &\sum_{k=1}^{K} w_k \sum_{i=1}^{N}\left\|\tilde{\boldsymbol{x}}_{gi}^k\boldsymbol{p_g}^k - y_i\right\|^2 \\ &+ w_{K+1}\sum_{i=1}^{N}\left\|\tilde{\boldsymbol{h}}_{gi}\boldsymbol{p_g}^{K+1} - y_i\right\|^2 + \lambda_1\sum_{k=1}^{K+1} w_k \ln w_k \\ &+\frac{1}{2}\lambda_2\sum_{k=1}^{K+1}\left(\boldsymbol{p_g}^k\right)^{\mathrm{T}}\boldsymbol{p_g}^k + \lambda_3\sum_{k=1}^{K}\sum_{i=1}^{N}\left\|\tilde{\boldsymbol{x}}_{gi}^k\boldsymbol{p_g}^k - \bar{y}_i^k\right\|^2 \\ &+\lambda_3\sum_{i=1}^{N}\left\|\tilde{\boldsymbol{h}}_{gi}\boldsymbol{p_g}^{K+1} - \bar{y}_i^{K+1}\right\|^2 \\ s.t.&\sum_{k=1}^{K+1} w_k = 1, w_k \geq 0\end{aligned} \tag{14}$$

where $y_i$ is the label of Sample *i*, $\boldsymbol{p_g}^k$ is the consequent parameter of the TSK FLS for the *k*th visible view, and $w_k$ is its corresponding weight. $\sum_{k=1}^{K}\sum_{i=1}^{N}\left\|\tilde{\boldsymbol{x}}_{gi}^k\boldsymbol{p_g}^k - \bar{y}_i^k\right\|^2$ and $\sum_{i=1}^{N}\left\|\tilde{\boldsymbol{h}}_{gi}\boldsymbol{p_g}^{k+1} - \bar{y}_i^{k+1}\right\|^2$ are the cooperation terms.

$\bar{y}_i^k = \frac{1}{K}\sum_{j=1,\ j\neq k}^{K+1} \tilde{\boldsymbol{x}}_{gi}^j \boldsymbol{p}_{\boldsymbol{g}}^j$ (for convenience, here $\tilde{\boldsymbol{x}}_{gi}^{K+1} = \tilde{\boldsymbol{h}}_{gi}$ ). $\sum_{k=1}^{K+1} w_k \ln w_k$ is the negative Shannon entropy corresponding to the weights of different views (including the hidden view), and $\sum_{k=1}^{K+1}\left(\boldsymbol{p}_{\boldsymbol{g}}^k\right)^{\mathrm{T}} \boldsymbol{p}_{\boldsymbol{g}}^k$ is a regularization term.

The roles of the terms in (14) are:

(i) The first two terms, $\sum_{k=1}^{K} w_k \sum_{i=1}^{N}\left\|\tilde{\boldsymbol{x}}_{gi}^k \boldsymbol{p}_{\boldsymbol{g}}^k - y_i\right\|^2$ and $w_{k+1}\sum_{i=1}^{N}\left\|\tilde{\boldsymbol{h}}_{gi}\boldsymbol{p}_{\boldsymbol{g}}^{K+1} - y_i\right\|^2$, are used to train the FLSs for multiple visible views and the hidden view.

(ii) The cooperation terms, $\sum_{k=1}^{K}\sum_{i=1}^{N}\left\|\tilde{\boldsymbol{x}}_{gi}^k \boldsymbol{p}_{\boldsymbol{g}}^k - \bar{y}_i^k\right\|^2$ and $\sum_{i=1}^{N}\left\|\boldsymbol{h}_{gi}\boldsymbol{p}_{\boldsymbol{g}}^{K+1} - \bar{y}_i^{K+1}\right\|^2$, try to make the outputs from different views consensus, which may help generalization.

(iii) The Shannon entropy term is used to determine the optimal weights for different views. Let $\sum_{k=1}^{K+1} w_k = 1,\ w_k \geq 0$. Then the weights can be regarded as a probability mass distribution, whose Shannon entropy is $-\sum_{i=1}^{K+1} w_k \ln w_k$ . Minimizing $\sum_{i=1}^{K+1} w_k \ln w_k$ will make $w_k$ close to each other, reducing the risk that a certain view dominates the final output.

(iv) The regularization parameters $\lambda_1 > 0$ , $\lambda_2 > 0$ , $\lambda_3 > 0$ are used to control the impact of the corresponding terms. They can be determined by CV.

### C. Parameter Learning

The Lagrangian method is used for solving $\boldsymbol{p}_{\boldsymbol{g}}^k$ and $w_k$ in (14), and the updating rules are:

$$\left(\boldsymbol{p}_{\boldsymbol{g}}^k\right)^{(t+1)} = \left[\lambda_2 \boldsymbol{I}_d + \left(w_k^{(t)} + \lambda_3\right)\sum_{i=1}^{N}\left(\tilde{\boldsymbol{x}}_{gi}^k\right)^{\mathrm{T}}\tilde{\boldsymbol{x}}_{gi}^k\right]^{-1} \cdot \left[w_k^{(t)}\sum_{i=1}^{N}\left(\tilde{\boldsymbol{x}}_{gi}^k\right)^{\mathrm{T}} y_i + \lambda_3 \sum_{i=1}^{K}\left(\tilde{\boldsymbol{x}}_{gi}^k\right)^{\mathrm{T}} (\bar{y}_i^k)^{(t)}\right] \tag{15}$$

$$w_k^{(t+1)} = \frac{\exp\left(-\sum_{i=1}^{N}\left(\tilde{\boldsymbol{x}}_{gi}^k\left(\boldsymbol{p}_{\boldsymbol{g}}^k\right)^{(t+1)} - y_i\right)\Big/\lambda_1\right)}{\sum_{k=1}^{K+1}\exp\left(-\sum_{i=1}^{N}\left(\tilde{\boldsymbol{x}}_{gi}^k\left(\boldsymbol{p}_{\boldsymbol{g}}^k\right)^{(t+1)} - y_i\right)\Big/\lambda_1\right)} \tag{16}$$

For convenience, we have set $\tilde{\boldsymbol{x}}_{gi}^{K+1} = \tilde{\boldsymbol{h}}_{gi}$ in (16).

Given a test sample with $K$ visible views, the output of the multi-view FLS is

$$\begin{aligned} Y_{output} &= \sum_{k=1}^{K+1} w_k f^k(\boldsymbol{x}) \\ &= \left(\sum_{k=1}^{K} w_k (\tilde{\boldsymbol{x}}_{\boldsymbol{g}}^k)^T \boldsymbol{p}_{\boldsymbol{g}}^k\right) + w_{K+1}\tilde{\boldsymbol{h}}_{\boldsymbol{g}}^T \boldsymbol{p}_{\boldsymbol{g}}^{K+1} \end{aligned} \tag{17}$$

$$f^k(\boldsymbol{x}) = \begin{cases} (\tilde{\boldsymbol{x}}_{\boldsymbol{g}}^k)^T \boldsymbol{p}_{\boldsymbol{g}}^k & 1 \leq k \leq K \\ \tilde{\boldsymbol{h}}_{\boldsymbol{g}}^T \boldsymbol{p}_{\boldsymbol{g}}^{K+1} & k = K+1 \end{cases} \tag{18}$$

### D. Generation of Hidden View Data of the Test Dataset

Given a test dataset with $N_{te}$ sample and $K$ views, the subset associated with the $k$th view can be represented as $\left\{\boldsymbol{x}_{\boldsymbol{i}}^{\boldsymbol{k}}\right\}_{i=1}^{N_{te}} \subset R^{d_k}$ , and the corresponding matrix form is $\bar{\boldsymbol{X}}_{te}^k = \left[\boldsymbol{x}_{\boldsymbol{1}}^{\boldsymbol{k}},\ldots,\boldsymbol{x}_{N_{te}}^{\boldsymbol{k}}\right]^T \in \mathrm{R}^{N_{te}\times d_k}$ . When $\bar{\boldsymbol{X}}_{te}^k$ has already been normalized to be non-negative, the hidden view data of the test data can be generated by optimizing the following objective function:

$$\begin{aligned} &\min_{H} \sum_{k=1}^{K}\left(\alpha^k \left\|\bar{\boldsymbol{X}}_{te}^k - \boldsymbol{H}\boldsymbol{W}^{*k}\right\|_F^2 + \beta^k Trace\left(\boldsymbol{H}^{\mathrm{T}}\boldsymbol{L}^k\boldsymbol{H}\right)\right) \\ &s.t.\ \boldsymbol{H} \geq 0 \end{aligned} \tag{19}$$

where $\boldsymbol{W}^{*k}$ is the mapping matrix of view $k$ and it is obtained based on the training data by solving (7). Since $\boldsymbol{W}^{*k}$ is fixed in (19), we only need to solve $\boldsymbol{H}$ with the following iterative rule:

$$\boldsymbol{H}_{i,j}^{(t+1)} \leftarrow \left[\frac{\sum_{k=1}^{K}\alpha^k\left(\bar{\boldsymbol{X}}_{te}^k\left(\boldsymbol{W}^{*k}\right)^T\right)_{i,j}}{\sum_{k=1}^{K}\left(\alpha^k\left(\boldsymbol{H}^{(t)}\boldsymbol{W}^{*k}\left(\boldsymbol{W}^{*k}\right)^T\right)_{i,j} + \beta^k\left(\boldsymbol{L}^k\boldsymbol{H}^{(t)}\right)_{i,j}\right)}\right]\boldsymbol{H}_{i,j}^{(t)} \tag{20}$$

Here, (20) is obtained with the similar way to (11).

### E. Theoretical Analysis for TSK-FLS-CVH

There exist two key problems in multi-view learning: 1) how to make full use of the otherness of different views, and 2) how to fully exploit the consistency among different views. The proposed TSK-FLS-CVH answers the above questions from the following two perspectives.

i) By introducing hidden view data, the proposed TSK-FLS-CVH can take advantage of the consistency information of different views. Because the hidden space is shared by different views, the hidden view data can reflect the internal relation between different views, i.e., the hidden view data contain the consistency information of the visible views. And to enhance the consistency among different views, the proposed method tries to make the outputs from different views consensus which may also help generalization.

ii) The proposed TSK-FLS-CVH exploits the otherness among different views by first training a TSK-FLS for each view data, including the hidden view data, respectively, and then assigning different weights to different views. It uses the maximum entropy in optimization to make the weights more sensible.

## IV. Experiments

Extensive experimental results are presented in this section to validate the performance of the proposed TSK-FLS-CVH. Five datasets were from the UCI machine learning repository and another is Epileptic EEG dataset [17]. The details are given in Table I. The performance index was the classification accuracy in five-fold cross validation.

Table I The multi-view datasets used in our experiments

| **Dataset** | **Number of Samples** | **Number of Classes** | **Visible View 1** | | **Visible View 2** | |
|---|---|---|---|---|---|---|
| | | | Description | Number of Features | Description | Number of Features |
| Forest Type | 326 | 4 | *Image band view*: Image band information of the data | 9 | *Spectrum view*: Spectrum values and the difference values | 18 |
| SPECTF | 267 | 2 | *Stress view*: Single proton emission computed tomography image of heart in stress | 22 | *Rest view*: Single proton emission computed tomography image of heart in rest | 22 |
| Dermatology | 366 | 6 | *Histopathological view*: Histopathological information of a case | 12 | *Clinical view*: Clinical information of a case | 22 |
| Multiple Features | 2000 | 10 | *Fourier coefficients view*: Fourier coefficients of the character shapes | 76 | *Zernike moments view*: Zernike moments of the character shapes | 47 |
| Image Segmentation | 2310 | 7 | *Shape view*: Shape information of an image | 9 | *RGB view*: RGB information of an image | 10 |
| Epileptic EEG | 500 | 2 | *DWT view: The features are extracted by DWT method as in [42].* | 20 | *WPD view: The features are extracted by WPD method. After decomposition of the EEG signals using WPD, nodes (1, 0), (2, 0), (3, 0), (4, 0), (5, 0) and (5, 1) in the binary tree were used to calculate the corresponding energies. A 6-dimensional feature vector was then generated for each signal.* | 6 |

## A. *Comparison with Single-view Algorithms*

The TSK-FLS-CVH was first compared with five single-view FLS algorithms: F-ELM [28], IQP [29], LESSLI [29], GENFIS [30] and L2-TSK-FS [31]. For the latter, we constructed a single-view dataset by combining features from all visible views.

In order to maintain the interpretability and compactness of the FLSs, the number of fuzzy rules for each single-view FLS was determined using grid search in $\{10, 12, 14, 16, 18, 20\}$. The average classification accuracies and the standard deviations of the six algorithms are shown in Table II. The proposed TSK-FLS-CVH achieved the highest accuracies in all six datasets, suggesting that multi-view learning is advantageous to single-view learning.

Table II Classification accuracies of the TSK-FLS-CVH and five single-view algorithms

| **Dataset** | TSK-FLS-CVH | F-ELM | IQP | LESSLI | GENFIS2 | L2-TSK-FS |
|---|---|---|---|---|---|---|
| Forest Type | **0.9086± 0.0206** | 0.8700± 0.0200 | 0.8127± 0.0246 | 0.8108± 0.0232 | 0.8451± 0.0048 | 0.8509± 0.0359 |
| SPECTF | **0.8688± 0.035** | 0.8203± 0.0306 | 0.7978± 0.0302 | 0.8240± 0.0391 | 0.7604± 0.0692 | 0.8164± 0.0250 |
| Dermatology | **0.9836± 0.0061** | 0.9016± 0.0264 | 0.8472± 0.0660 | 0.8715± 0.0669 | 0.9699± 0.0115 | 0.9781± 0.0184 |
| Multiple Feature | **0.8735± 0.0136** | 0.6230± 0.0418 | 0.4955± 0.0229 | 0.4580± 0.0265 | 0.8695± 0.0096 | 0.8405± 0.0130 |
| Image Segmentation | **0.8879± 0.0136** | 0.8675± 0.0412 | 0.7970± 0.0247 | 0.6970± 0.0192 | 0.8468± 0.0135 | 0.6476± 0.1037 |
| Epileptic EEG | **0.9700±0.0084** | 0.9460±0.0305 | 0.9680±0.011 | 0.9640±0.0114 | 0.9640±0.0167 | 0.8960±0.0182 |
| Mean | **0.9154** | 0.8381 | 0.7864 | 0.7709 | 0.8760 | 0.8382 |

## B. *Comparison with Multi-view Algorithms*

The TSK-FLS-CVH was next compared with two multi-view algorithms, TwoV-TSK-FCS [16] and AMVMED [32]. The latter two are only available for binary classification. We used a 1-v-1 strategy to extend them to multi-class classification.

We also transformed the five single-view algorithms in the previous subsection to the corresponding multi-view algorithms: a single-view classifier was trained separately for each visible view, and then their outputs were averaged. To distinguish from the original single-view algorithms, the transformed algorithms are denoted as F-ELM (MV), IQP (MV), LESSLI (MV), GENFIS2 (MV) and L2-TSK-FS (MV), respectively.

The experimental results on different multi-view datasets are shown in Tables III-VIII.

Table III Classification accuracies (Mean ± Std) of the eight multi-view algorithms on the Forest Type dataset

| Algorithm | Forest type | | | |
|---|---|---|---|---|
| | Image Band View | Spectrum View | Hidden View | Integration of Multiple Views |
| TSK-FLS-CVH | 0.8872 ± 0.0139 | 0.8816 ± 0.0309 | 0.7686 ± 0.0182 | **0.9086 ± 0.0206** |
| TwoV-TSK-FCS | 0.8604 ± 0.0106 | 0.6578 ± 0.0345 | N/A | 0.8604 ± 0.0106 |
| AMVMED | 0.8853 ± 0.0064 | 0.8873 ± 0.0293 | N/A | 0.9026 ± 0.0243 |
| F-ELM (MV) | 0.8834 ± 0.0104 | 0.8413 ± 0.0148 | N/A | 0.8872 ± 0.0139 |
| IQP (MV) | 0.8108 ± 0.0284 | 0.7859 ± 0.0329 | N/A | 0.8203 ± 0.0262 |
| LESSLI (MV) | 0.8050 ± 0.0357 | 0.7668 ± 0.0377 | N/A | 0.8146 ± 0.0310 |
| GENFIS2 (MV) | 0.8489 ± 0.0327 | 0.8624 ± 0.0169 | N/A | 0.8643 ± 0.0227 |
| L2-TSK-FS (MV) | 0.8298 ± 0.0239 | 0.6272 ± 0.0844 | N/A | 0.7783 ± 0.0483 |

Table IV Classification accuracies (Mean ± Std) of the eight multi-view algorithms on the SPECTF dataset

| Algorithm | SPECTF | | | |
|---|---|---|---|---|
| | Stress View | Rest View | Hidden View | Integration of Multiple Views |
| TSK-FLS-CVH | 0.8015 ± 0.0172 | 0.8203 ± 0.0275 | 0.8462 ± 0.0416 | **0.8688 ± 0.0350** |
| TwoV-TSK-FCS | 0.7940 ± 0.0021 | 0.6029 ± 0.0544 | N/A | 0.8277 ± 0.0277 |
| AMVMED | 0.7940 ± 0.0021 | 0.8015 ± 0.0161 | N/A | 0.8015 ± 0.0161 |
| F-ELM (MV) | 0.7753 ± 0.0290 | 0.8167 ± 0.0415 | N/A | 0.8092 ± 0.0460 |
| IQP (MV) | 0.7867 ± 0.0398 | 0.7679 ± 0.0419 | N/A | 0.8017 ± 0.0397 |
| LESSLI (MV) | 0.7904 ± 0.0325 | 0.7979 ± 0.0227 | N/A | 0.8166 ± 0.0355 |
| GENFIS2 (MV) | 0.7978 ± 0.0153 | 0.7754 ± 0.0212 | N/A | 0.7940 ± 0.0021 |
| L2-TSK-FS (MV) | 0.7940 ± 0.0021 | 0.8052 ± 0.0117 | N/A | 0.8090 ± 0.0242 |

Table V Classification accuracies (Mean ± Std) of the eight multi-view algorithms on the Dermatology dataset

| Algorithm | Dermatology | | | |
|---|---|---|---|---|
| | Clinical View | Histopathological View | Hidden View | Integration of Multiple Views |
| TSK-FLS-CVH | 0.8660 ± 0.0396 | 0.9454 ± 0.0194 | 0.9563 ± 0.0062 | 0.9836 ± 0.0061 |
| TwoV-TSK-FCS | 0.4507 ± 0.0509 | 0.7979 ± 0.0305 | N/A | 0.8715 ± 0.0538 |
| AMVMED | 0.8032 ± 0.0539 | 0.9344 ± 0.1103 | N/A | 0.9699 ± 0.0179 |
| F-ELM (MV) | 0.7649 ± 0.0435 | 0.8906 ± 0.0680 | N/A | **0.9841 ± 0.0179** |
| IQP (MV) | 0.6612 ± 0.0571 | 0.8177 ± 0.0557 | N/A | 0.7405 ± 0.0393 |
| LESSLI (MV) | 0.6448 ± 0.0291 | 0.8034 ± 0.0635 | N/A | 0.7760 ± 0.0108 |
| GENFIS2 (MV) | 0.8496 ± 0.0392 | 0.9399 ± 0.0155 | N/A | 0.9672 ± 0.0076 |
| L2-TSK-FS (MV) | 0.7895 ± 0.0582 | 0.9426 ± 0.0225 | N/A | 0.9699 ± 0.0225 |

Table VI Classification accuracies (Mean ± Std) of the eight multi-view algorithms on the Multiple Features dataset

| Algorithm | Multiple Features | | | |
|---|---|---|---|---|
| | Fourier Coefficients View | Zemike View | Hidden View | Integration of Multiple Views |
| TSK-FLS-CVH | 0.8180 ± 0.0226 | 0.8215 ± 0.0060 | 0.7475 ± 0.0095 | **0.8735 ± 0.0191** |
| TwoV-TSK-FCS | 0.5765 ± 0.0201 | 0.6370 ± 0.0310 | N/A | 0.7225 ± 0.0225 |
| AMVMED | 0.4300 ± 0.0419 | 0.7995 ± 0.0114 | N/A | 0.8660 ± 0.0096 |
| F-ELM (MV) | 0.5525 ± 0.0326 | 0.6340 ± 0.0552 | N/A | 0.7375 ± 0.0326 |
| IQP (MV) | 0.3690 ± 0.0194 | 0.4050 ± 0.0232 | N/A | 0.4500 ± 0.0208 |
| LESSLI (MV) | 0.3120 ± 0.0259 | 0.3615 ± 0.0464 | N/A | 0.3490 ± 0.0460 |
| GENFIS2 (MV) | 0.7825 ± 0.0170 | 0.8060 ± 0.0080 | N/A | 0.8430 ± 0.0149 |
| L2-TSK-FS (MV) | 0.7180 ± 0.0251 | 0.7035 ± 0.0251 | N/A | 0.7990 ± 0.0154 |

Table VII Classification accuracies (Mean ± Std) of the eight multi-view algorithms on the Image Segmentation dataset

| Algorithm | Image Segmentation | | | |
|---|---|---|---|---|
| | Shape View | RGB View | Hidden View | Integration of Multiple Views |
| TSK-FLS-CVH | 0.8015 ± 0.0172 | 0.8203 ± 0.0275 | 0.8462 ± 0.0416 | 0.8879 ± 0.0136 |
| TwoV-TSK-FCS | 0.3870 ± 0.0179 | 0.7009 ± 0.0213 | N/A | 0.7009 ± 0.0213 |
| AMVMED | 0.7009 ± 0.0213 | 0.7740 ± 0.0104 | N/A | **0.9364 ± 0.0129** |
| F-ELM (MV) | 0.4701 ± 0.0865 | 0.8134 ± 0.0148 | N/A | 0.8771 ± 0.0146 |
| IQP (MV) | 0.4320 ± 0.0073 | 0.6065 ± 0.0199 | N/A | 0.5455 ± 0.0136 |
| LESSLI (MV) | 0.3939 ± 0.0363 | 0.6273 ± 0.0513 | N/A | 0.4662 ± 0.0529 |
| GENFIS2 (MV) | 0.4853 ± 0.0182 | 0.7342 ± 0.0210 | N/A | 0.8429 ± 0.0172 |
| L2-TSK-FS (MV) | 0.3693 ± 0.0238 | 0.6745 ± 0.0200 | N/A | 0.6710 ± 0.0766 |

Table VIII Classification accuracies (Mean ± Std) of the eight multi-view algorithms on the Epileptic EEG dataset

| Algorithm | Epileptic EEG dataset. | | | |
|---|---|---|---|---|
| | WAV View | WPD View | Hidden View | Integration of Multiple Views |
| TSK-FLS-CVH | 0.9360 ± 0.0207 | 0.9480± 0.0130 | 0.7320 ± 0.0130 | **0.9700 ± 0.0084** |
| TwoV-TSK-FCS | 0.6620± 0.0740 | 0.9080± 0.0179 | N/A | 0.9240 ± 0.0182 |
| AMVMED | 0.6520 ± 0.0327 | 0.8640± 0.0297 | N/A | 0.9500± 0.0381 |
| F-ELM (MV) | 0.7660 ± 0.1062 | 0.9080± 0.0164 | N/A | 0.9540 ± 0.0114 |
| IQP (MV) | 0.9600 ± 0.0187 | 0.9480 ± 0.0130 | N/A | **0.9700 ± 0.0158** |
| LESSLI (MV) | 0.9540 ± 0.0241 | 0.9440 ± 0.0195 | N/A | **0.9700 ± 0.0148** |
| GENFIS2 (MV) | 0.9260± 0.0114 | 0.9300 ± 0.0158 | N/A | 0.9440 ± 0.0130 |
| L2-TSK-FS (MV) | 0.6760 ± 0.0564 | 0.8800± 0.0354 | N/A | 0.9180 ± 0.0259 |

The proposed TSK-FLS-CVH obtained the best classification accuracies in four out of the six multi-view datasets. On some datasets, the classification accuracies of some multi-view algorithms were worse than those obtained from a single-view. For example, Table V shows that on the Dermatology dataset, the classification accuracy of IQP (MV) when using only the histopathological view was 0.8177, but its multi-view learning accuracy was only 0.7405. This suggests that a simple average integration of multiple views may not always enhance the performance. However, TSK-FLS-CVH, which employs a more sophisticated strategy to integrate different views, can always achieve better performance than each individual view.

In particular, although the performance of the proposed method is better than many existing related methods on the binary classification dataset SPECTF, its advantage seems not obvious as that in some multi-class datasets. It may be due to the fact that multi-class classification is usually more complicated than binary classification. In this situation, the proposed method will be more adaptive than the other related methods due to its comprehensive cooperative learning abilities from both the visible views and the shared hidden view simultaneously.

### *C. Statistical Analysis*

*1) Friedman test combined with post-Holm test:* Friedman test [33, 34] was used to check if there was statistically significant difference among the 13 algorithms. The null hypothesis, which says that the classification performances of all methods were the same, was rejected ($p$=0.0023), i.e., the classification performances of the 13 algorithms had statistically significant difference. Table IX shows the ranking of the 13 algorithms. A lower ranking indicates a better performance. Clearly, TSK-FLS-CVH had the best performance.

A post-Holm test was then used to compare the best method, i.e., TSK-FLS-CVH with the other 12 algorithms. The $p$-value and the statistical magnitude are shown in Table X. The null hypothesis, i.e., there does not exist statistically significant difference between two algorithms, is rejected if $p < 0.05$. Table X shows that there was statistically significant difference between TSK-FLS-CVH and L2-TSK-FS, L2-TSK-FS (MV), IQP, IQP (MV), LESSLI, LESSLI (MV), GENFIS2, GENFIS2 (MV), F-ELM and TwoV-TSK-FCS, respectively. Although the null hypothesis was not rejected when compared with F-ELM (MV) and AMVMED, Tables II-VIII show that TSK-FLS-CVH still outperformed them slightly.

*2) t-test*: *t-test* based on the classification accuracies of five-fold cross validation on different dataset was used to further compare the proposed method with others. The results are shown in Table XI. The null hypothesis, i.e., there does not exist statistically significant difference between two algorithms, is rejected if $p < 0.05$, and the "Rejected numbers" index is the number of datasets on which the hypothesis is rejected. Table XI shows that the proposed method significantly outperformed GENFIS2, GENFIS2(MV), L2-TSK-FS, L2-TSK-FS(MV), IQP, IQP(MV), LESSLI, LESSLI(MV), and TwoV-TSK-FCS on most datasets (no less than half of the number of datasets). Although the null hypothesis was not rejected on most datasets when compared with F-ELM, F-ELM(MV) and AMVMEND, the proposed method still outperformed them slightly, as shown in Tables II-VIII.

Table IX Ranking of the 13 algorithms based on Friedman test

| Algorithm | Ranking |
|---|---|
| TSK-FLS-CVH | 1.333 |
| AMVMED | 4.6667 |
| F-ELM(MV) | 5.6667 |
| GENFIS2 | 6.25 |
| F-ELM | 6.3333 |
| GENFIS2(MV) | 7.1667 |
| L2-TSK-FS | 7.3333 |
| TwoV-TSK-FCS | 7.4167 |
| LESSLI | 8.3333 |
| L2-TSK-FS(MV) | 8.8333 |
| IQP | 9 |
| LESSLI(MV) | 9.1667 |
| IQP(MV) | 9.5 |

Table X The results of Post Holm test

| **Algorithm** | $z = (R_0 - R_i) / SE$ | $p$ | $\text{Holm} = \alpha / i$, $\alpha = 0.05$ | **Null Hypothesis** |
|---|---|---|---|---|
| TwoV-TSK-FCS/ TSK-FLS-CVH | 3.632122 | 0.000281 | 0.004167 | Rejected |
| GENFIS2(MV)/ TSK-FLS-CVH | 3.483872 | 0.000494 | 0.004545 | Rejected |
| L2-TSK-FS(MV)/ TSK-FLS-CVH | 3.409747 | 0.00065 | 0.005 | Rejected |
| IQP(MV)/ TSK-FLS-CVH | 3.335622 | 0.000851 | 0.005556 | Rejected |
| LESSLI(MV)/ TSK-FLS-CVH | 3.113247 | 0.00185 | 0.00625 | Rejected |
| IQP/ TSK-FLS-CVH | 2.70556 | 0.006819 | 0.007143 | Rejected |
| F-ELM(MV)/ TSK-FLS-CVH | 2.668498 | 0.007619 | 0.008333 | Rejected |
| LESSLI TSK-FLS-CVH | 2.594373 | 0.009476 | 0.01 | Rejected |
| GENFIS2/ TSK-FLS-CVH | 2.223748 | 0.026165 | 0.0125 | Rejected |
| L2-TSK-FS/ TSK-FLS-CVH | 2.186685 | 0.028765 | 0.016667 | Rejected |
| F-ELM/ TSK-FLS-CVH | 1.927248 | 0.053949 | 0.025 | Not rejected |
| AMVMED/ TSK-FLS-CVH | 1.482499 | 0.138208 | 0.05 | Not rejected |

Table XI The results （*p value*） of *t*-test

| **Algorithm** | **Dataset** | | | | | | **Rejected Numbers** |
|---|---|---|---|---|---|---|---|
| | Forest Type | SPECTF | Dermatology | Multiple Feature | Image Segmentation | Epileptic EEG | |
| TwoV-TSK-FCS/ TSK-FLS-CVH | 0.0183 | 0.2506 | 0.0016 | 0.0001 | 0.0031 | 0.0026 | 5 |
| L2-TSK-FS(MV)/ TSK-FLS-CVH | 0.0005 | 0.1008 | 0.2829 | 0.0067 | 0.0002 | 0.0032 | 4 |
| IQP(MV)/ TSK-FLS-CVH | 0.0003 | 0.0911 | 0 | 0 | 0 | 1 | 4 |
| LESSLI(MV)/ TSK-FLS-CVH | 0.0043 | 0.1699 | 0 | 0 | 0 | 1 | 4 |
| IQP/TSK-FLS-CVH | 0.0001 | 0.0536 | 0.0021 | 0 | 0.0012 | 0.8974 | 4 |
| LESSLI/ TSK-FLS-CVH | 0.0003 | 0.3804 | 0.0055 | 0 | 0 | 0.6395 | 4 |
| GENFIS2(MV)/ TSK-FLS-CVH | 0.0116 | 0.0127 | 0.3972 | 0.8363 | 0.0386 | 0.7485 | 3 |
| GENFIS2/ TSK-FLS-CVH | 0.0116 | 0.0127 | 0.3972 | 0.8363 | 0.0386 | 0.7485 | 3 |
| L2-TSK-FS/ TSK-FLS-CVH | 0.0032 | 0.1508 | 0.7637 | 0.2097 | 0.0025 | 0.0002 | 3 |
| F-ELM(MV) / TSK-FLS-CVH | 0.0994 | 0.1402 | 0.0815 | 0.0006 | 0.3744 | 0.2543 | 2 |
| F-ELM/ TSK-FLS-CVH | 0.0766 | 1 | 0.0002 | 0 | 0.5077 | 0.1143 | 2 |
| AMVMED/ TSK-FLS-CVH | 0.691 | 0.0847 | 0.3356 | 0.717 | 0.007 | 0.2914 | 1 |

Table XII Classification accuracy of TSK-FLS-CVH with and without the hidden view

| **Dataset** | **TSK-FLS-CVH (without the hidden view)** | **TSK-FLS-CVH (with the hidden view)** |
|---|---|---|
| Forest Type | 0.8949 ± 0.0132 | **0.9086** ± 0.0206 |
| SPECTF | 0.8054 ± 0.0196 | **0.8688** ± 0.0350 |
| Dermatology | 0.9754 ± 0.0114 | **0.9836** ± 0.0061 |
| Multiple Features | 0.8480 ± 0.0135 | **0.8735** ± 0.0136 |
| Image Segmentation | **0.9078** ± 0.0094 | 0.8879 ± 0.0136 |
| EEG Singals | 0.9560±0.0089 | **0.9700 ± 0.0084** |
| Average | 0.8979 | **0.9154** |

## *D. Effect of the Hidden View*

The effect of the hidden view is further studied in this subsection.

First, to test whether the hidden view can indeed enhance the multi-view learning performance, we compared the classification accuracies of TSK-FLS-CVH with and without the hidden view. The results are shown in Table XII. Observe that the hidden view helped increase the classification accuracies on most datasets.

The dimensionality of the hidden view is an adjustable parameter. Its effect is also studied. Let the minimum dimensionality of all visible views be $r$. We extracted 9 hidden view datasets with dimensionalities $\lceil 0.1*r \rceil$, $\lceil 0.2*r \rceil$, ..., $\lceil 0.9*r \rceil$ ( $\lceil\ \rceil$ denotes the round operation), respectively. The regularization parameters $\lambda_1, \lambda_2, \lambda_3$ in TSK-FLS-CVH were optimized without using the hidden view, and then hidden views with different dimensionalities were introduced. Experimental results on Forest Type, SPECT and Dermatology are shown in Fig. 5, which indicate that the dimensionality of the hidden view can slightly influence the classification accuracy. How to optimally set this parameter is still an open problem. In our experiment, hidden views with different dimensionalities were first trained, and the one with the best classification accuracy was then used in the subsequent cooperative learning between the visible and hidden views.

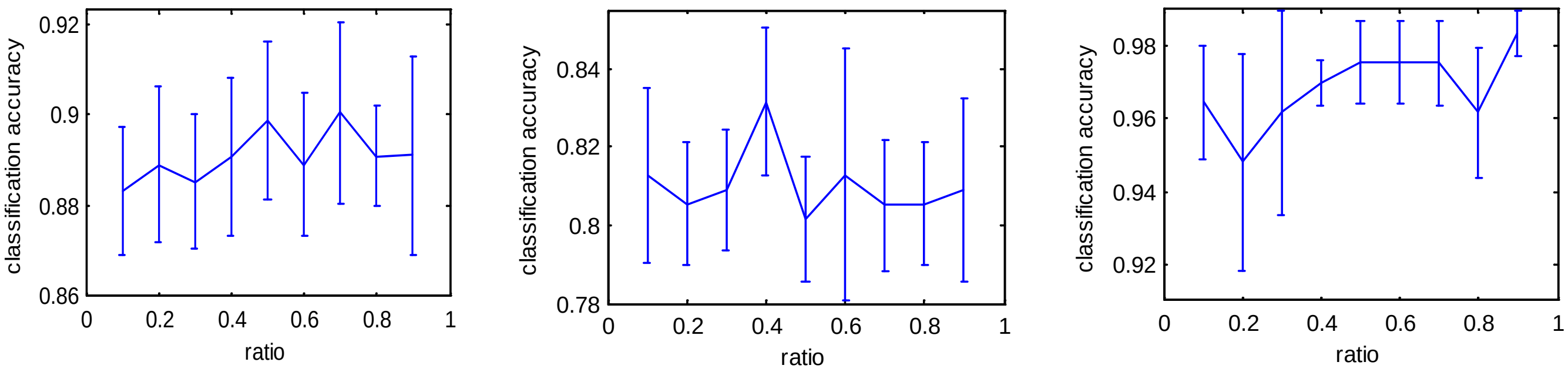

(a) Forest type (b) SPECTF (c) Dermatology

Fig. 5 The influence of the dimensionality of the hidden view on the classification accuracy.

### *E. An Example of The Generated Multi-view FLSs*

In this section, an example of the generated multi-view TSK FLS by the proposed TSK-FLS-CVH based on the Forest Type dataset is shown. In order to readily display the results, we set the number of fuzzy rules of FLSs in visible and hidden views to 3 and 4，respectively. Due to page limit, more results are presented in the Supplementary materials. In Tables S1 and S2 of the Supplementary materials, the parameters of the fuzzy rules of the FLSs generated in the image band view and the shared hidden view are listed, respectively. Meanwhile, the membership functions of the fuzzy subsets in fuzzy rules and the corresponding linguistic descriptions are shown for two views in Figs. S1 and S2, respectively. Based on these results, the first fuzzy rule in the two FLSs of two views can be linguistically described as follows.

The first fuzzy rule of the TSK FLS generated in the image band view is:

*IF the band 1 of ASTER image is Low,*
*and the band 2 of ASTER image is Medium,*
*and the band 3 of ASTER image is Medium,*
*and the band 4 of ASTER image is Low,*
*and the band 5 of ASTER image is Medium,*
*and the band 6 of ASTER image is Medium,*
*and the band 7 of ASTER image is Low,*
*and the band 8 of ASTER image is Medium,*
*And the band 9 of ASTER image is Low.*
*THEN this rule gives decision value of the first output with the following formula:*

$$f^2(\boldsymbol{x}) = \begin{bmatrix} 0.0066 + 0.0372x_1 - 0.0620x_2 \\ -0.0263x_3 - 0.0034x_4 - 0.0215x_5 + \\ 0.0143x_6 - 0.0345x_7 - 0.0550x_8 + 0.0588x_9 \end{bmatrix}$$

The first fuzzy rule of the TSK FLS generated in the shared hidden view is:

*IF the shared hidden feature 1 is High,*
*and the shared hidden feature 2 is Low,*
*and the shared hidden feature 3 is High,*
*and the shared hidden feature 4 is A little low,*
*and the shared hidden feature 5 is High,*
*and the shared hidden feature 6 is Low,*
*and the shared hidden feature 7 is A Little High,*
*THEN this rule gives decision value of the first output with the following formula:*

$$f^1(\boldsymbol{x}) = \begin{bmatrix} -0.2350 - 0.2496x_1 - 0.2348x_2 - 0.2229x_3 \\ -0.2293x_4 - 0.2660x_5 - 0.2241x_6 - 0.4652x_7 \end{bmatrix}$$

Finally, the decision of the multi-view FLS for the first output can be given as follows.

$$f(\boldsymbol{x}) = w_1 f(\boldsymbol{x})_{\text{image-band-view}} + w_2 f(\boldsymbol{x})_{\text{spectrum-view}} \\ w_3 f(\boldsymbol{x})_{\text{shared-hidden-view}} .$$

## V. CONCLUSIONS

In this paper, a multi-view FLS with the cooperation between visible and hidden views has been proposed. It uses not only the information from multiple visible views, but also the information of the shared hidden view among multiple visible views, to enhance the learning performance. Experimental results demonstrated that the proposed approach outperformed many traditional single-view approaches and some state-of-the-art multi-view approaches.

Our future research includes how to efficiently optimize the hyper-parameters, and how to optimally determine the dimensionality of the hidden view.

# Supplementary Materials of the Manuscript

## "Multi-View Fuzzy Logic System with the Cooperation between Visible and Hidden Views"

Table S1 The parameters of the fuzzy rules of the FLSs generated in the image band view by the proposed TSK-FLS-CVH method based on the Forest Type dataset

| **Fuzzy rules base** | | | |
|---|---|---|---|
| TSK Fuzzy Rule $R^k$: | | | |
| $IF\ x_1\ is\ A_1^k\left(c_1^k,\delta_1^k\right)\wedge x_1\ is\ A_1^k\left(c_1^k,\delta_1^k\right)\wedge\ldots\wedge x_1\ is\ A_1^k\left(c_1^k,\delta_1^k\right),\ THEN\ f_k\left(x\right)=p_{k0}+p_{k1}x_1+\ldots+p_{kd}x_d$ | | | |
| **View** | **No. of rules** | **Antecedent parameters** (Gaussian membership function parameters) | **Consequent parameters** (linear function parameters) |
| *Image band* view | $k$ | $\boldsymbol{c}^k=\left(c_1^k,\ldots,c_d^k\right)^T,\delta^k=\left(\delta_1^k,\ldots,\delta_d^k\right)^T$ | $\boldsymbol{p}_k=\left(p_{k0},p_{k1},\ldots,p_{kd}\right)^T$ |
| | 1 | $c^1=[0.2534,0.1054,0.0863,0.3350,0.260$ $0.2981,0.4383,0.1004,0.1573]$<br>$\delta^1=[0.0226,0.0089,0.0081,0.0090,0.02$ $0.0187,0.0210,0.0082,0.0104]$ | $\boldsymbol{p}_1=[1.0766,-1.3624,-0.6184,$ $0.0259,-0.1573,-0.5457,$ $-0.6371,-0.0618,0.2182,0.7179]$ |
| | 2 | $c^2=[0.4806,0.335,0.2838,0.4514,0.5844,$ $0.6104,0.6997,0.3007,0.4368]$<br>$\delta^2=[0.0237,0.0226,0.0195,0.0169,0.04$ $0.0443,0.0476,0.0261,0.0341]$ | $\boldsymbol{p}_2=[0.0066,0.0372,-0.0620,$ $-0.0263,-0.0034,-0.0215,$ $0.0143,-0.0345,-0.0550,0.0588]$ |
| | 3 | $c^3=[0.6401,0.0478,0.0533,0.5150,0.137$ $0.2296,0.5958,0.0976,0.2188]$<br>$\delta^3=[0.0292,0.0077,0.0061,0.0112,0.01$ $0.0142,0.0161,0.0064,0.0068]$ | $\boldsymbol{p}_3=[-0.0094,-0.1547,-1.0403,$ $-0.9756,-0.1148,-0.0937,$ $-0.2083,0.4227,0.7971,0.6919]$ |

Table S2 The parameters of the fuzzy rules of the FLS generated in the shared hidden view by the proposed TSK-FLS-CVH method based on the Forest Type dataset

| **Fuzzy rules base** | | | |
|---|---|---|---|
| TSK Fuzzy Rule $R^k$: | | | |
| $IF\ x_1\ is\ A_1^k\left(c_1^k,\delta_1^k\right)\wedge x_1\ is\ A_1^k\left(c_1^k,\delta_1^k\right)\wedge...\wedge x_1\ is\ A_1^k\left(c_1^k,\delta_1^k\right),\ THEN\ f_k\left(x\right)=p_{k0}+p_{k1}x_1+...+p_{kd}x_d$ | | | |
| **View** | **No. of rules** | **Antecedent parameters** (Gaussian membership function parameters) | **Consequent parameters** (linear function parameters) |
| *Shared hidden* view | *k* | $\boldsymbol{c}^k=\left(c_1^k,...,c_d^k\right)^T,\delta^k=\left(\delta_1^k,...,\delta_d^k\right)^T$ | $\boldsymbol{p}_k=\left(p_{k0},p_{k1},...,p_{kd}\right)^T$ |
| | 1 | $c^1=[0.4596,0.1138,0.5000,0.1760,$ $0.5000,0.0000,0.3645]$ <br> $\delta^1=[0.0967,0.1000,0.0233,0.0610,$ $0.0534,0.0529,0.0155]$ | $\boldsymbol{p}_1=[-0.23498,-0.2496,-0.2348,$ $-0.2229,-0.2293,-0.2660,$ $-0.2241,-0.4652]$ |
| | 2 | $c^2=[0.2433,0.2753,0.2724,0.1852,$ $0.2013,0.1752,0.3259]$ <br> $\delta^2=[0.0488,0.0462,0.0345,0.0407,$ $0.0397,0.0358,0.0264]$ | $\boldsymbol{p}_2=[-0.31646,-0.3935,-0.3195,$ $-0.4696,-0.3086,-0.2402,$ $-0.5522,2.0000]$ |
| | 3 | $c^3=[0.0657,0.4322,0.2036,0.1156,$ $0.0653,0.1918,0.3884]$ <br> $\delta^3=[0.0127,0.0130,0.0234,0.0141,$ $0.0170,0.0129,0.0175]$ | $\boldsymbol{p}_3=[-0.20182,-0.1895,-0.0902,$ $-0.1389,-0.0656,-0.4868,$ $0.1512,-0.4250]$ |
| | 4 | $c^4=[0.2943,0.2154,0.2337,0.2477,$ $0.1741,0.2261,0.2701]$ <br> $\delta^4=[0.0573,0.0518,0.0458,0.0523,$ $0.0510,0.0465,0.0374]$ | $\boldsymbol{p}_4=[-0.2519,-0.2594,-0.2112,$ $-0.1017,-0.1656,-0.0119,$ $-0.3301,-1.1171]$ |

| | ***Rule 1*** | ***Rule 2*** | ***Rule 3*** |
|---|---|---|---|
| Band1 | 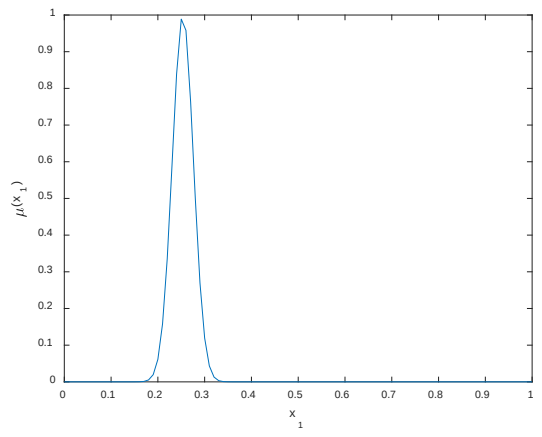<br>(**0.25**,0.02)*<br>***Low*****| 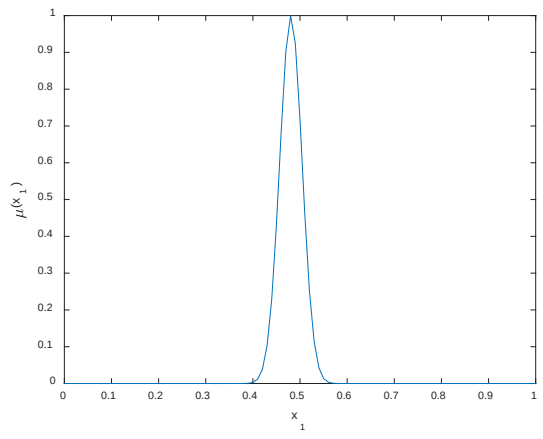<br>(**0.48**,0.02)<br>***Medium*** | 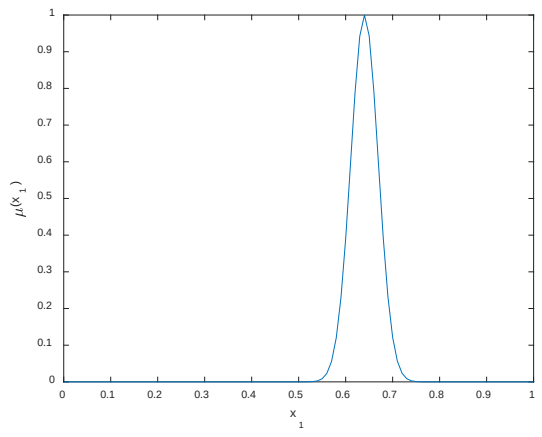<br>(**0.64**,0.03)<br>***High*** |
| Band2 | 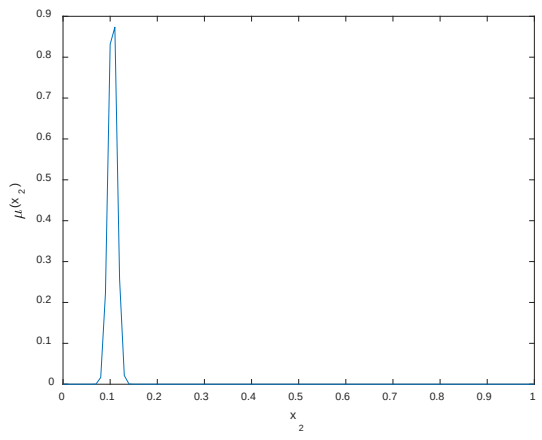<br>(**0.11**,0.01)<br>***Medium*** | 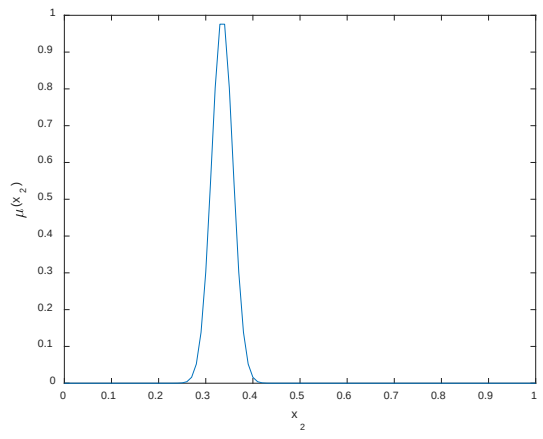<br>(**0.33**,0.02)<br>***High*** | 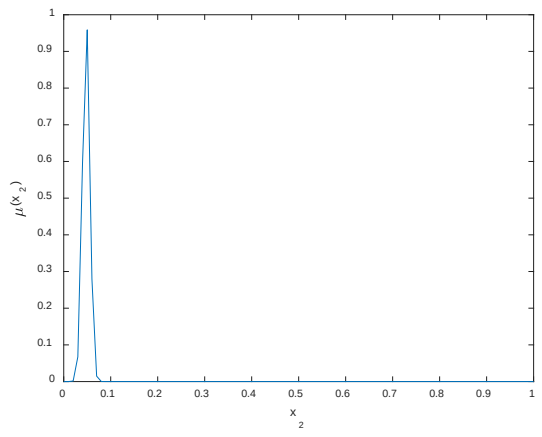<br>(**0.05**,0.01)<br>***Low*** |
| Band3 | 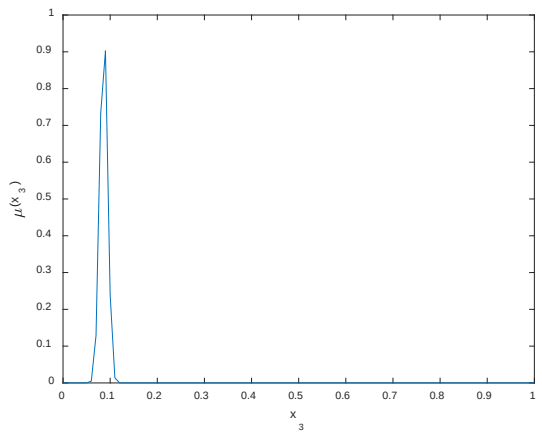<br>(**0.09**,0.01)<br>***Medium*** | 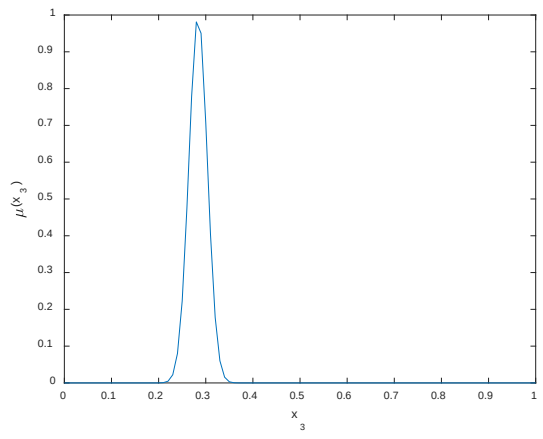<br>(**0.28**,0.02)<br>***High*** | 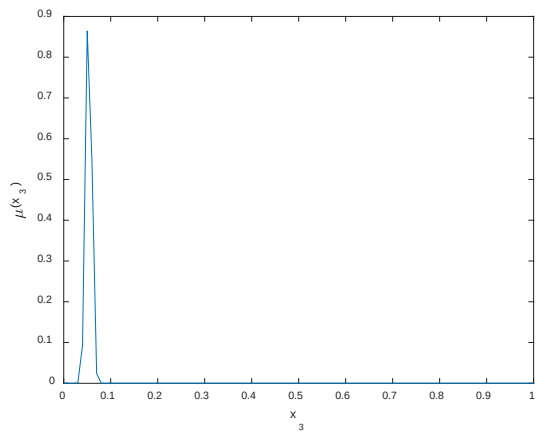<br>(**0.05**,0.01)<br>***Low*** |
| Band4 | 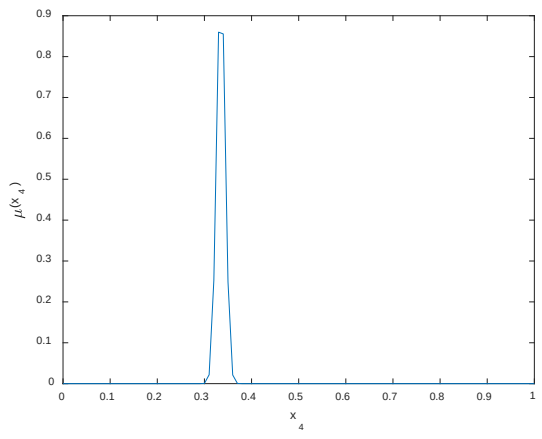<br>(**0.33**,0.01)<br>***Low*** | 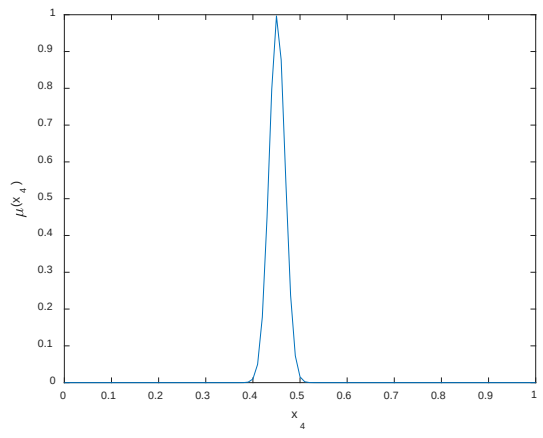<br>(**0.45**,0.02)<br>***Medium*** | 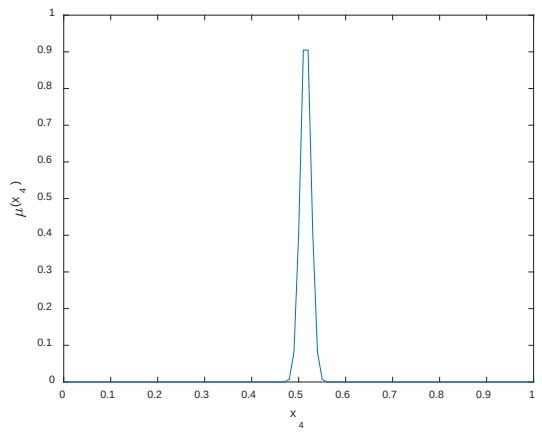<br>(**0.51**,0.01)<br>***High*** |
| Band5 | 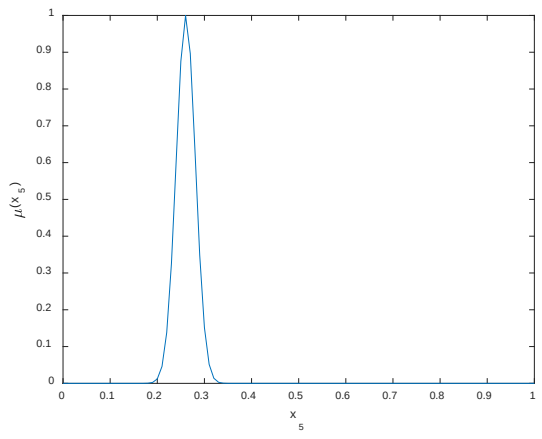<br>(**0.26**,0.02)<br>***Medium*** | 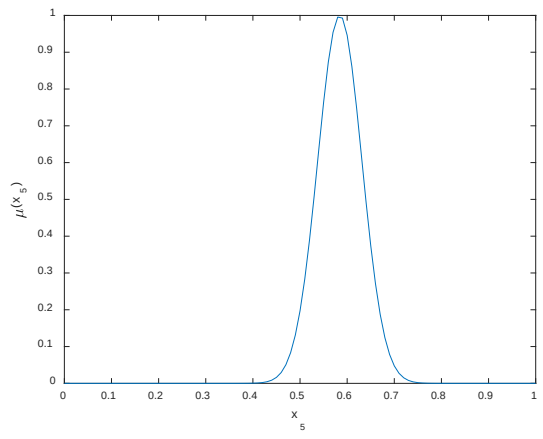<br>(**0.58**,0.05)<br>***High*** | 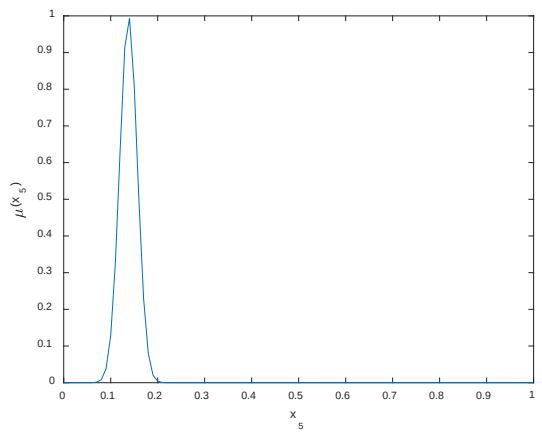<br>(**0.14**,0.02)<br>***Low*** |

Band6

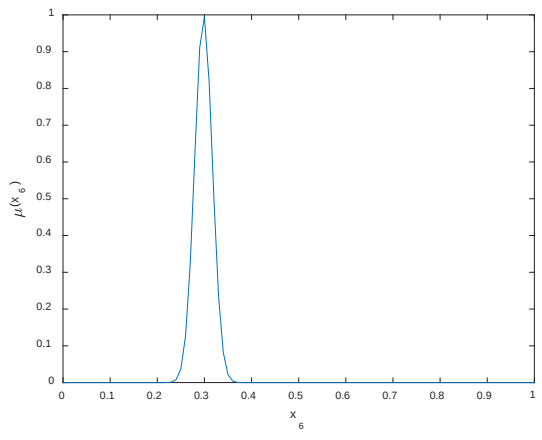

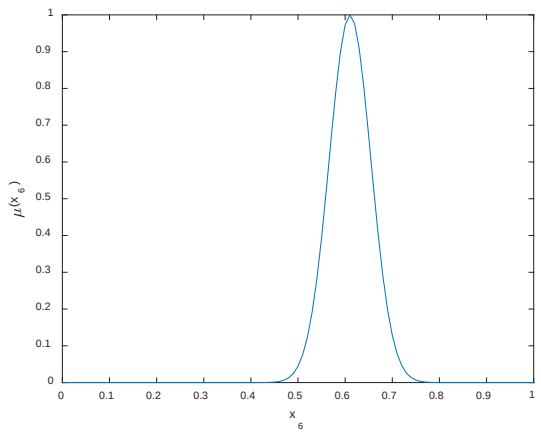

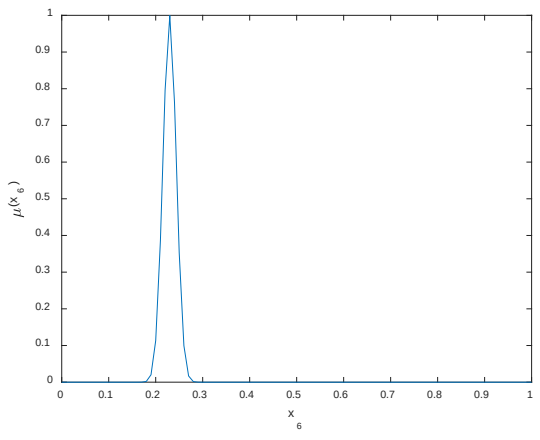

(**0.30**,0.02) (**0.61**,0.04) (**0.23**,0.01)

***Medium*** ***High*** ***Low***

Band7

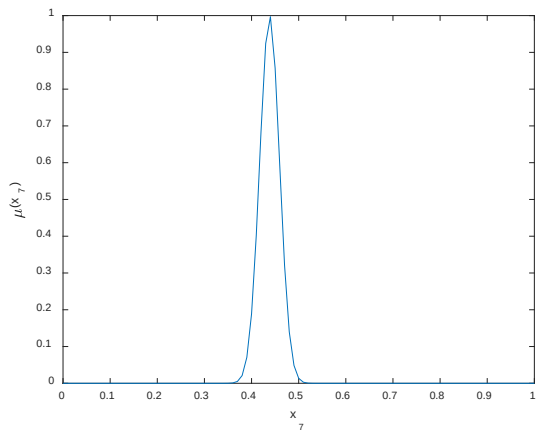

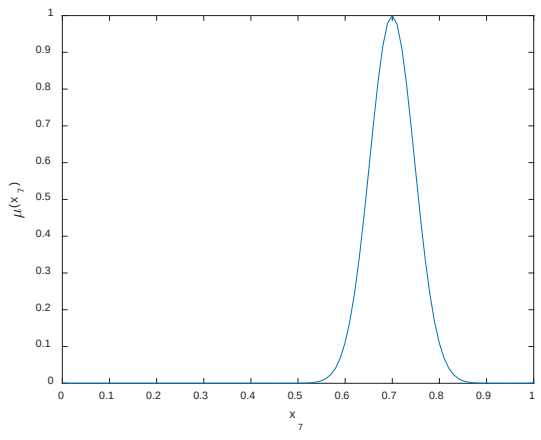

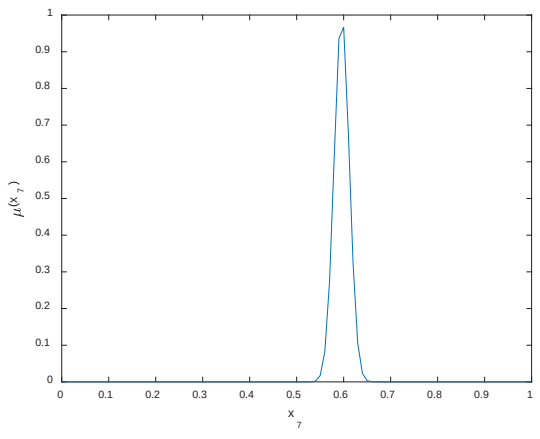

(**0.44**,0.02) (**0.70**,0.05) (**0.60**,0.02)

***Low*** ***High*** ***Medium***

Bnad8

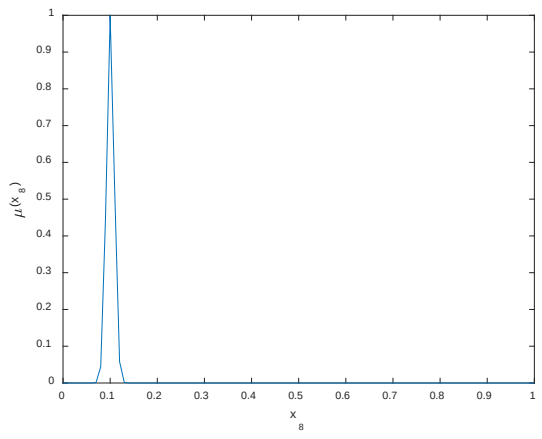

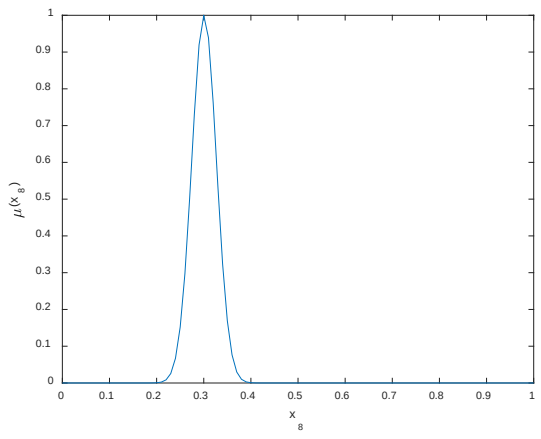

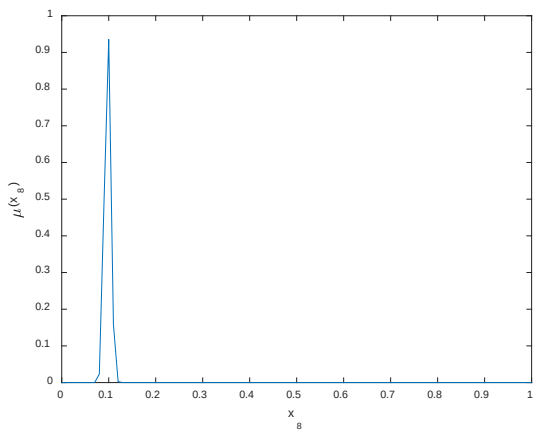

(**0.10**,0.01) (**0.30**,0.03) (**0.09**,0.01)

***Medium*** ***High*** ***Low***

Band9

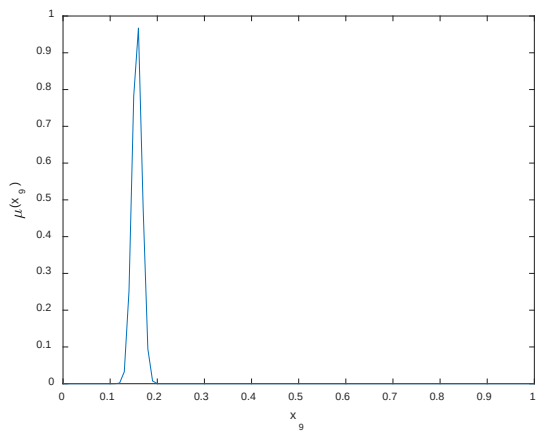

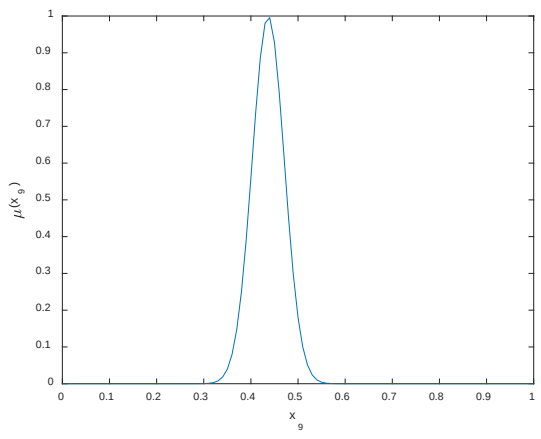

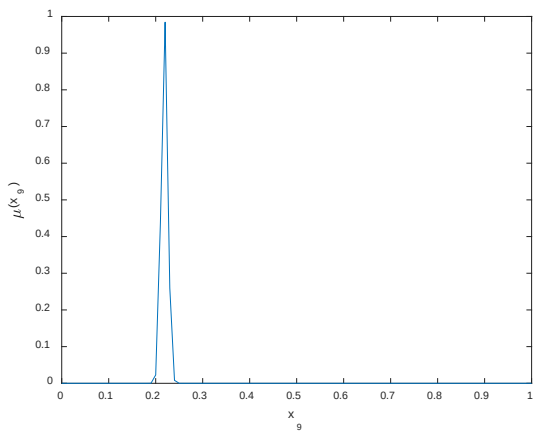

(**0.16**,0.01) (**0.44**,0.03) (**0.22**,0.01)

***Low*** ***High*** ***Medium***

Fig. S1 Membership functions and the potential linguistic explanation of each fuzzy subset in the antecedent of each fuzzy rule of the TSK FLS obtained by the proposed TSK-FLS-CVH method for the Image band view.

* Antecedent parameter of band 1(first dimension of data) of first fuzzy rule.

** Potential explanation for the fuzzy set obtained.

| | *Rule 1* | *Rule 2* | *Rule 3* | *Rule 4* |
|---|---|---|---|---|
| Hidden Feature 1 | 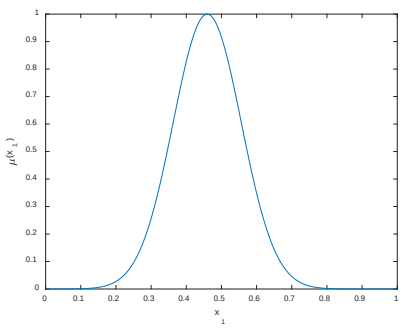 (**0.46**,0.10)[*] *High*[**] | 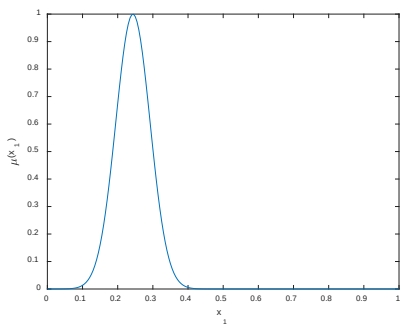 (**0.24**,0.05) *A little low* | 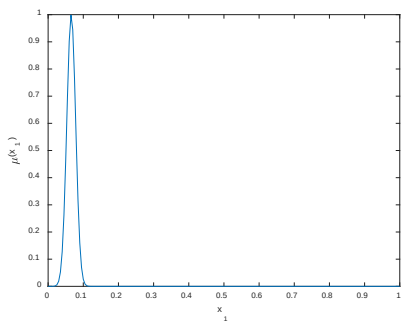 (**0.07**,0.01) *Low* | 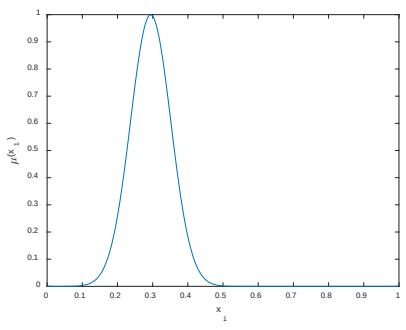 (**0.29**,0.06) *A little high* |
| Hidden Feature 2 | 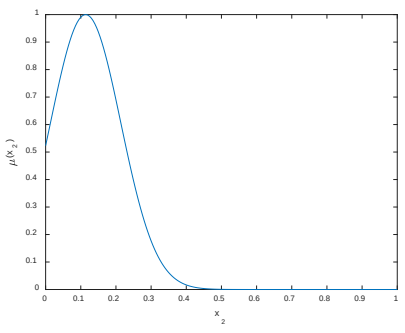 (**0.11**,0.10) *Low* | 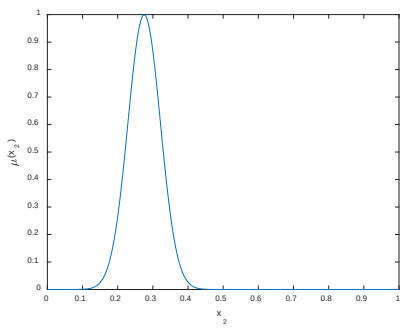 (**0.28**,0.05) *A little high* | 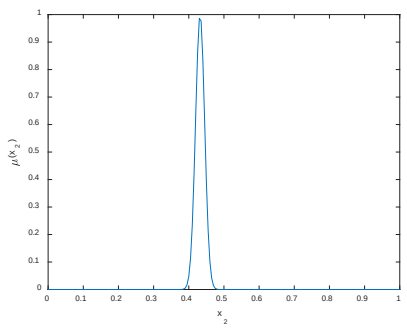 (**0.43**,0.01) *High* | 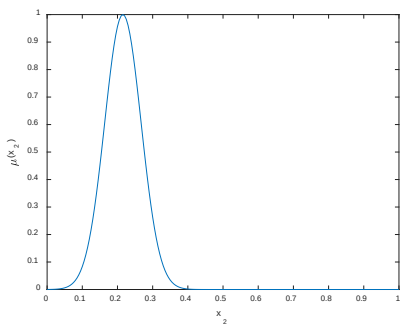 (**0.22**,0.05) *A little low* |
| Hidden Feature 3 | 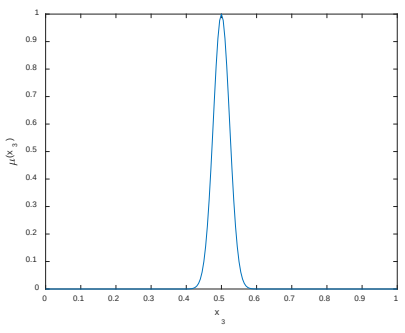 (**0.50**,0.02) *High* | 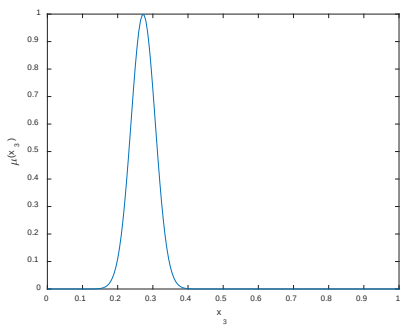 (**0.27**,0.03) *A little high* | 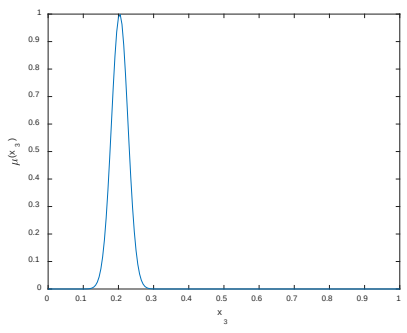 (**0.20**,0.02) *Low* | 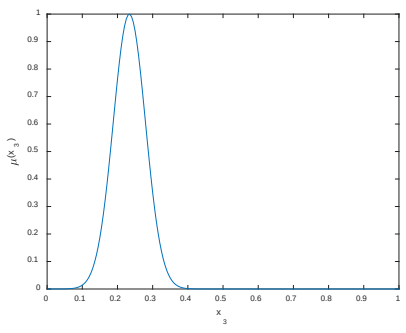 (**0.23**,0.05) *A little low* |
| Hidden Feature 4 | 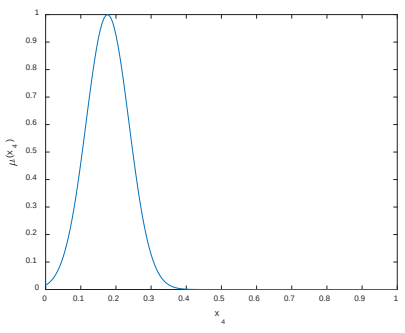 (**0.18**,0.06) *A little low* | 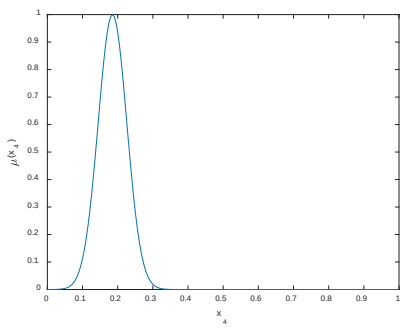 (**0.19**,0.04) *A little high* | 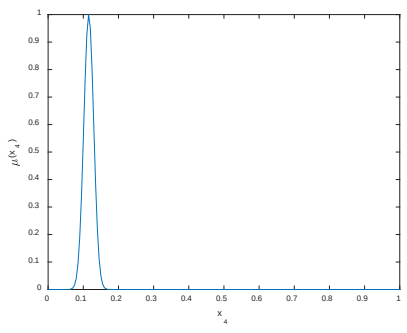 (**0.12**,0.01) *Low* | 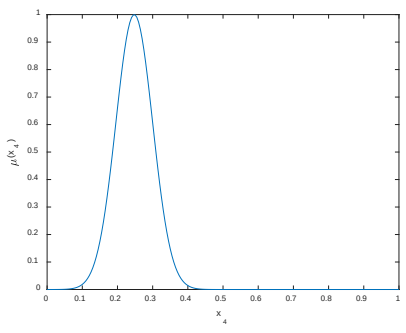 (**0.25**,0.05) *High* |
| Hidden Feature 5 | 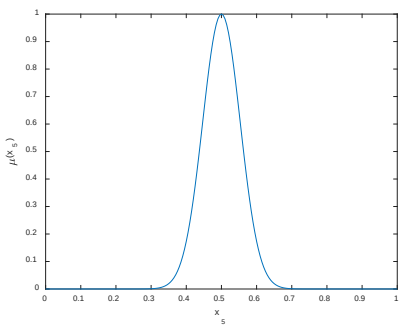 (**0.50**,0.05) *High* | 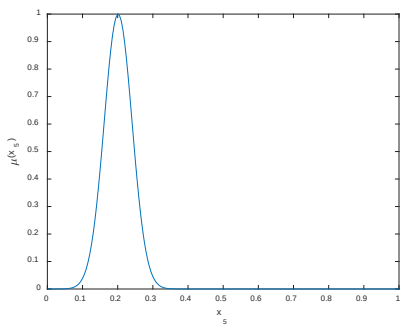 (**0.20**,0.04) *A little high* | 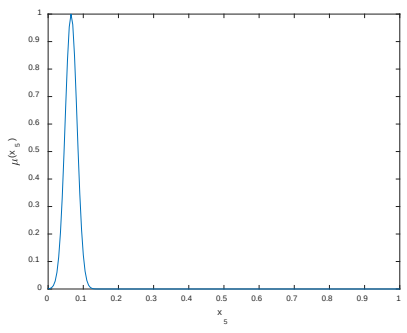 (**0.07**,0.02) *Low* | 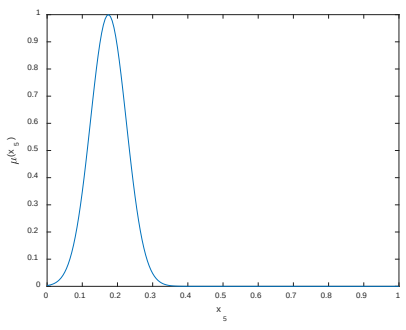 (**0.17**,0.05) A little low |
| Hidden Feature 6 | 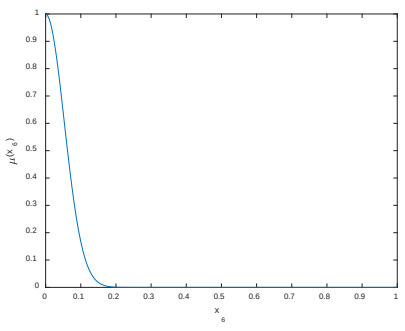 (**0**,0.05) *Low* | 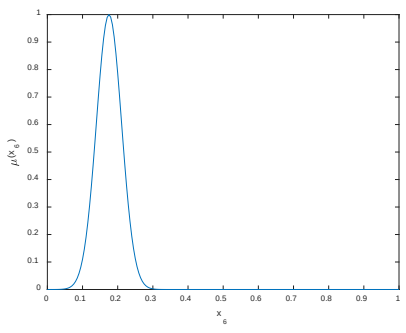 (**0.18**,0.04) *A little low* | 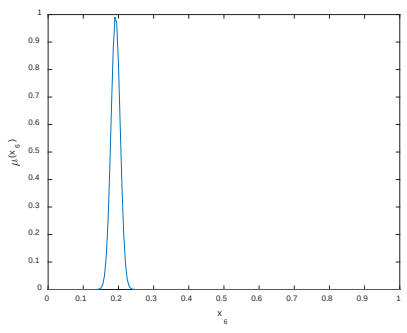 (**0.19**,0.01) *A little high* | 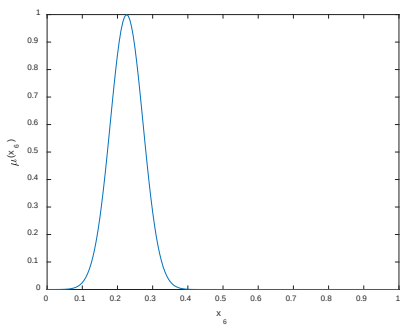 (**0.23**,0.05) *High* |

Hidden Feature 7

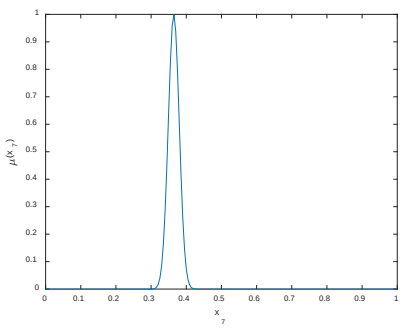

(**0.36**,0.02)
***A little high***

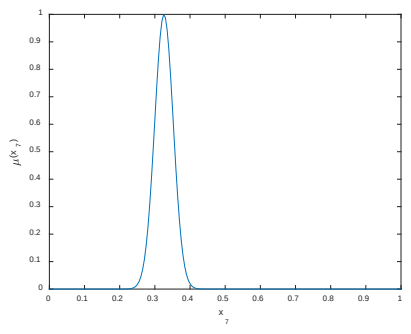

(**0.33**,0.03)
***A little low***

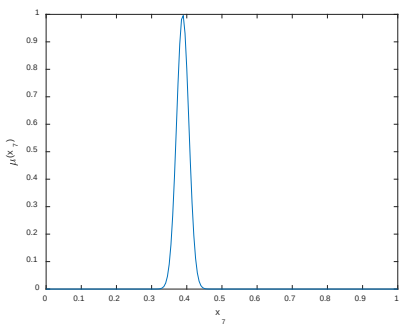

(**0.39**,0.03)
***High***

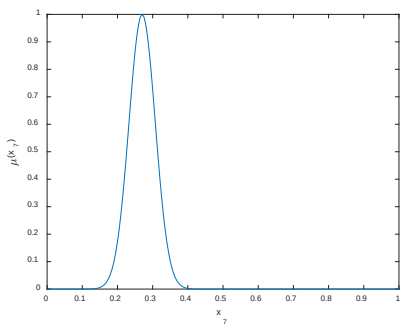

(**0.27**,0.04)
***Low***

Fig. S2 Membership functions and the potential linguistic explanation of each fuzzy subset in the antecedent of each fuzzy rule of the TSK FLS obtained by the proposed TSK-FLS-CVH method for the shared hidden view.

* Antecedent parameter of band 1(first dimension of data) of first fuzzy rule.

** Potential explanation for the fuzzy set obtained.